\documentclass[runningheads]{llncs}

\usepackage{eccv}

\usepackage{eccvabbrv}

\usepackage{graphicx}
\usepackage{booktabs}

\usepackage[accsupp]{axessibility}  
\definecolor{cvprblue}{rgb}{0.21,0.49,0.74}
\usepackage[pagebackref,breaklinks,colorlinks,allcolors=cvprblue]{hyperref}
\usepackage[ruled,vlined]{algorithm2e}
\usepackage{bm}
\usepackage{titletoc}
\newcommand{\authcount}[1]{}
\usepackage{authblk}
\usepackage[table]{xcolor}
\usepackage{graphicx}
\usepackage{comment}
\usepackage{multirow}

\definecolor{lightergray}{gray}{0.9}
\usepackage[most]{tcolorbox}
\tcbuselibrary{listings,breakable}
\usepackage{listings}
\usepackage{xcolor}

\newcommand{\styledfileinput}[3][python]{%
  \begin{tcolorbox}[
      enhanced,
      breakable,
      listing only,
      colback=gray!10,
      colframe=gray!10,
      left=0mm,right=0mm,top=0mm,bottom=0mm,
      listing options={language=#1,#3}
  ]
    \lstinputlisting{#2}
  \end{tcolorbox}
}

\usepackage{hyperref}

\usepackage{orcidlink}

\makeatletter
\renewcommand\subsubsection{\@startsection{subsubsection}{3}{\z@}%
  {-1.5ex\@plus -0.5ex \@minus -.2ex}%
  {0.8ex \@plus .2ex}%
  {\normalfont\normalsize\bfseries}}
\makeatother
\begin{document}

\title{Global Graph-Validated Optimization for VLM-based 3D Indoor Scene Generation}

\titlerunning{Graph-Validated 3D Scene Generation}

\author{
Jialu Huang\inst{1}\orcidlink{0009-0009-4112-2833} \
Yingxuan You\inst{2}\orcidlink{0000-0002-0154-4590} \
Fei Wang\inst{1}\orcidlink{0000-0003-3462-8472} \
Zheng Dang\inst{3}\thanks{Corresponding author.}\orcidlink{0000-0003-2028-6096}
}

\authorrunning{J.~Huang et al.}

\institute{
National Key Laboratory of Human-Machine Hybrid Augmented Intelligence \\
Institute of Artificial Intelligence and Robotics, Xi’an Jiaotong University, China
\and
École Polytechnique Fédérale de Lausanne (EPFL), CVLab, Switzerland
\and
School of Electronics and Information, Northwestern Polytechnical University and \\
Shaanxi Key Laboratory of Information Acquisition and Processing, China
\\
\email{\{huangjialu@stu,wfx@mail\}.xjtu.edu.cn} \\
\email{yingxuan.you@epfl.ch} \\
\email{zheng.dang@nwpu.edu.cn}
}

\maketitle
\begin{abstract}
We study open-vocabulary 3D indoor layout generation, which synthesizes diverse and physically plausible indoor scenes from unlabeled 3D assets given free-form language instructions.
Recent text-guided layout generation methods leverage large language models (LLMs) and vision-language models (VLMs) to synthesize structured scenes directly from text descriptions. 
However, most existing approaches model inter-asset relations implicitly or rely on local pairwise constraints during local optimization. 
Such formulations are misaligned with the global and highly non-convex feasible layout space, often producing locally plausible but globally inconsistent or physically infeasible scenes.
To address these limitations, we introduce a graph-based intermediate representation that decouples semantic coherence and physical feasibility, and propose a hybrid search-and-refinement strategy to generate indoor layouts with global semantic consistency and physical feasibility.
First, we propose Global Semantic Verification (GSV), which represents scenes as structured scene graphs and enforces semantic constraints through rule-based graph verification. This explicit structural validation prunes contradictory configurations and produces a globally consistent semantic scaffold for scene generation. 
Second, we introduce Global Physical Feasibility Search (GPFS), a hybrid optimization framework that combines evolutionary search for global exploration with gradient-based refinement for local exploitation. GPFS reduces dependence on VLM-proposed initialization and improves robustness in highly non-convex and discontinuous feasible spaces.
Together, GSV and GPFS shift layout generation from local relational modeling and initialization-sensitive optimization toward globally consistent reasoning and exploration. 
Experiments demonstrate that our method achieves state-of-the-art performance on open-vocabulary 3D indoor layout generation, improving both semantic consistency and physical plausibility.

\end{abstract}
   
\section{Introduction}

\begin{figure}[tp]
\centering
\includegraphics[width=0.95\linewidth]{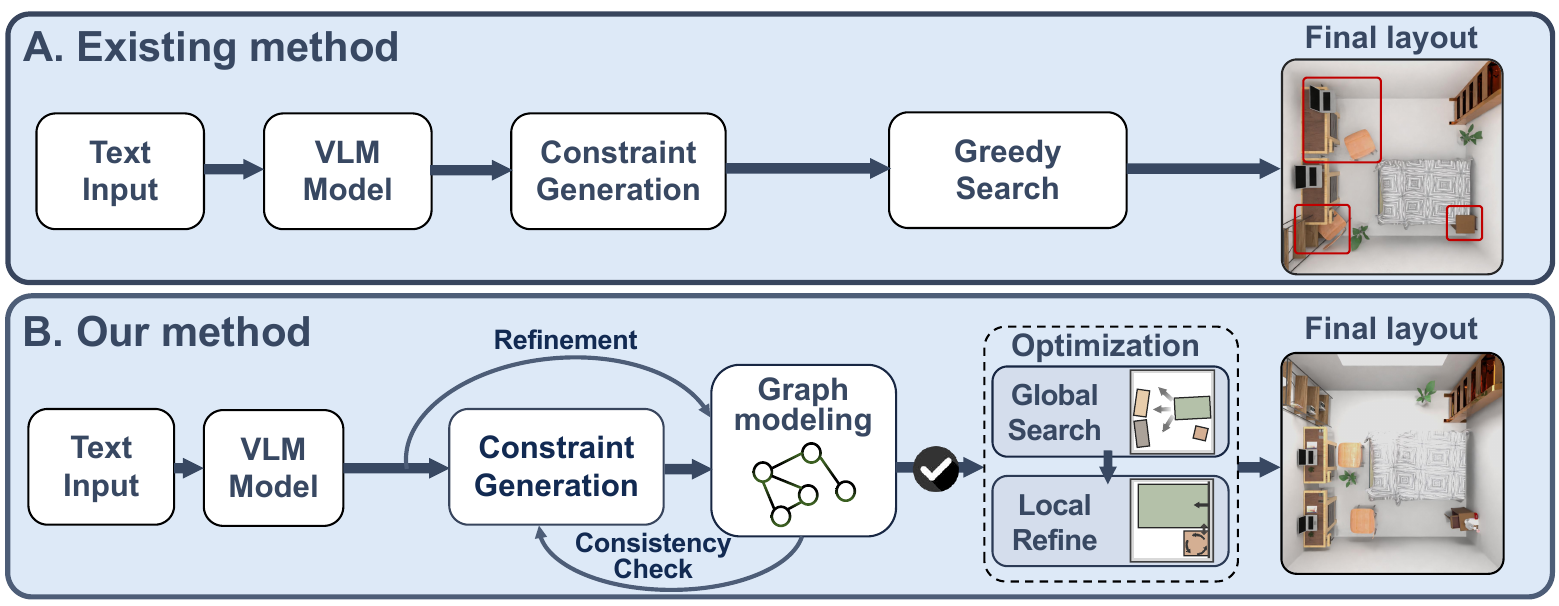}
\vspace{-0.2cm}
\caption{\textbf{Existing methods vs. our approach.} Prior methods rely on local pairwise consistency and gradient-based refinement, often producing semantically or physically invalid layouts. Our graph-based iterative refinement and optimization yields coherent and physically plausible layouts.}
\vspace{-0.4cm}
\label{fig:motivation}
\end{figure}


Spatial reasoning and planning refer to understanding, arranging, and manipulating objects in 3D space while respecting physical constraints. These abilities are crucial for autonomous agents to navigate, plan, and interact with complex environments. Automatically generating realistic simulated scenes has become essential for providing diverse training data to improve agents' spatial reasoning. In this work, we focus on open-vocabulary 3D indoor layout generation, creating diverse indoor scenes from unlabeled 3D assets guided solely by free-form language instructions.

Recent progress in text-guided 3D indoor layout generation leverages large language models (LLMs)~\cite{Feng23, Ccelen24, Yang24, Fu24, Deng25, Ling25} and vision-language models (VLMs)~\cite{Sun25} to synthesize structured scenes directly from text descriptions.
These methods implicitly model inter-object relations through probabilistic likelihood estimation, and some further enforce pairwise consistency between object pairs during optimization. 
However, scene layout generation is inherently a globally constrained reasoning problem. The validity of a scene depends not only on local pairwise compatibility, but on the joint satisfaction of semantic and physical constraints across all objects. 
This misalignment manifests in two critical limitations.
Firstly, existing VLM-based layout methods either model relations implicitly or rely on local pairwise regularization, lacking an explicit mechanism to ensure global consistency. 
As a result, scenes can be locally plausible but globally invalid.
Secondly, physical feasibility depends critically on the optimization strategy used to realize layouts. 
Recent methods predominantly rely on gradient-based optimization initialized from VLM predictions.
However, the feasible layout space under semantic and physical constraints is highly non-convex and often exhibits discontinuities due to collision avoidance, support relations, and stability requirements. Gradient descent is inherently local and strongly dependent on initialization, making it prone to being trapped in suboptimal basins. 
Consequently, the optimization process may converge to configurations that either violate physical constraints or fail to explore alternative feasible arrangements that better satisfy both semantic and physical requirements.

In this work, we address these two challenges by explicitly disentangling semantic coherence and physical feasibility, and introducing dedicated mechanisms for each.
To enforce global semantic coherence, we introduce the Global Semantic Verification (GSV). Instead of promoting consistency through probabilistic alignment, GSV models the scene as a structured scene graph and encodes semantic constraints explicitly. We then perform rule-based graph verification to check relational consistency and prune contradictory configurations.
This shifts layout generation from implicit relational fitting to explicit structural validation, producing a scene-level semantically consistent graph that serves as the scaffold for subsequent layout realization.

To improve physical feasibility and mitigate sensitivity to VLM initialization, we propose Global Physical Feasibility Search (GPFS) based on consistent graph generated by GSV for layout generation. 
GPFS combines evolutionary search for global exploration with gradient-based refinement for local exploitation. 
The evolutionary component maintains a diverse population of candidate layouts within the semantically constrained space, reducing reliance on a single starting point and helping escape local minima.
Notably, GPFS does not require a VLM-proposed initialization and can operate from a diverse population of randomly initialized layouts. 
The gradient component then refines candidate solutions to better satisfy semantic and physical constraints. 
Together, our GSV and GPFS move layout generation beyond local consistency checks and neighborhood refinement, enabling globally consistent exploration.
This shift from local to global reasoning ensures scene-level coherence and physical validity, improving the structural quality of automatically generated scenes.

Our contribution can be summarized as follows:
\begin{itemize}
\item We identify and disentangle two key limitations in VLM-based layout generation: the lack of global semantic coherence and the sensitivity of gradient-based optimization to physical feasibility constraints. 
    
\item We propose GSV, a rule-based graph verification mechanism that enforces global semantic coherence through explicit structural validation over scene graphs. To our knowledge, this is the first work to incorporate explicit scene graph modeling and verification into a VLM-based 3D indoor layout generation pipeline.
    
\item We introduce GPFS, a hybrid search framework that integrates evolutionary global exploration with gradient-based local refinement, reducing reliance on VLM initialization and improving reliability to highly non-convex and discontinuous feasible spaces.

\item We achieve state-of-the-art performance on open-vocabulary 3D indoor layout generation, with clear improvements in semantic consistency and physical plausibility.
\end{itemize}

\section{Related Work}

\noindent\textbf{Open-Vocabulary Indoor Scene Synthesis.}
Indoor layout generation and scene synthesis have historically relied on manually specified rules and hand-crafted statistical priors~\cite{Merrell11, Yeh12, Fisher12, Chang14, Chang17, Fu17}. With the availability of large-scale 3D scene datasets~\cite{Fu21a, Khanna24}, recent work~\cite{Li19b, Dhamo21, Wang21c, Hu24, Lin24, Tang24, Sun24, Wu24, Yang24a} began to learn composition patterns directly from data, marking a transition from early rule-based approaches to deep learning-driven methods. More recently, the emergence of large language models and vision-language models has enabled open-vocabulary 3D scene synthesis, allowing flexible layout generation guided by natural language instructions.

\noindent\textbf{Language-Guided Indoor Layout Generation.}
To improve flexibility beyond fixed object vocabularies and predefined scene templates, recent work has increasingly incorporated large language models (LLMs) and vision–language models (VLMs) into 3D layout generation pipelines, using natural language as a high-level specification for scene structure. LayoutGPT\cite{Feng23} pioneered this direction by employing GPT to generate room layouts from high-level instructions. Subsequent works such as~\cite{Aguina24, Ccelen24, Yang24, Fu24, Deng25, Ling25, Littlefair25, Su25} have also integrated LLMs into the synthesis pipeline. Further advancing this line, LayoutVLM\cite{Sun25} employs a vision-language model to define semantic relationships and uses differentiable optimization to produce continuous and semantically consistent layouts. 
However, they often struggle in scenes with many objects, the relationships generated by LLM become unreliable. In contrast, our method mitigates these limitations by improving the reliability of relational information and enabling robust optimization from scratch, yielding layouts that remain coherent and physically valid.

\noindent\textbf{Scene Graph Representation.}
Scene graphs, which represent objects as nodes and relationships as edges, have been widely adopted for structured 3D scene representation and support various tasks including scene retrieval~\cite{Wald20}, comparison~\cite{Fisher11}, and layout generation~\cite{Luo20}. This representation has been extended to 3D environments through hierarchical structures~\cite{Liu14, Armeni19, Li19b, Zhao11, Wald20}, parse trees~\cite{Purkait20}, and graph neural networks~\cite{Johnson18}. Within these approaches, generative methods often incorporate VAEs~\cite{Kingma13, Sohn15} and use message passing for layout synthesis~\cite{Zhou19b}. Recent methods, including Holodeck~\cite{Yang24}, InstructScene~\cite{Lin24}, and Graph-to-3D~\cite{Dhamo21}, continue this tradition by leveraging scene graphs for semantic control, instruction-driven editing, and geometry-aware generation.
However, these methods generally depend on a static scene graph that assumes well-formed inputs. In contrast, our approach maintains an adaptive graph that is refined during generation, providing more reliable relational cues to guide the VLM.

\section{Methodology}

\noindent\textbf{Problem Definition.}
We formulate the task as arranging a set of 3D assets within a bounded environment according to natural language instructions. Given a textual layout description $\ell_{\text{layout}}$, the system identifies a set of all semantically relevant 3D assets $\mathcal{A}=\{a_i\}^N_{i=1}$, where $N$ is the number of assets and each $a_i$ denotes a 3D asset instance inferred from the text. The objective is to estimate the pose of each asset:
\begin{equation}
P_i = (x_i, y_i, z_i, \theta_i),
\label{eq:pose}
\end{equation}
where $(x_i, y_i, z_i)$ represents the 3D position and $\theta_i$ denotes the orientation around the vertical axis. 
The final output is the scene configuration: $\mathcal{S} = \{(a_i, P_i)\}_{i=1}^N,$ assigning each asset a specific pose.

\begin{figure*}[!tp]
    \centering
    \includegraphics[width=\textwidth]{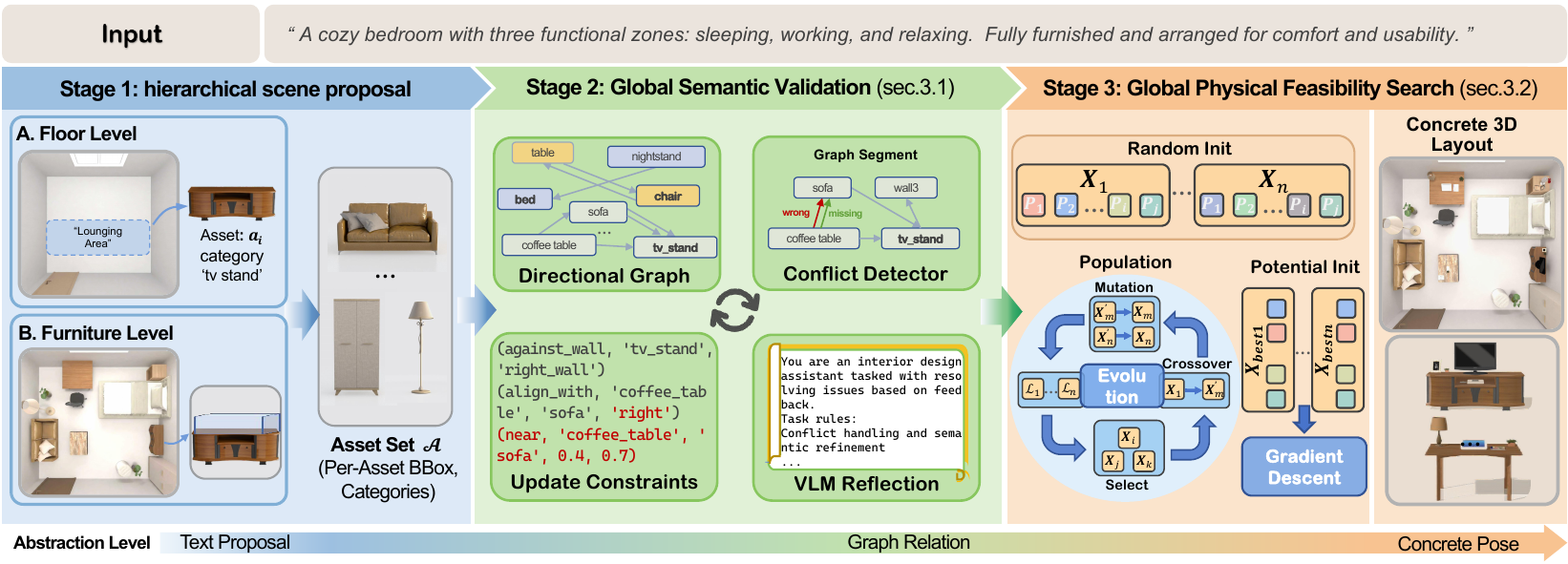}
    \vspace{-0.6cm}
    \caption{\textbf{Overview of the proposed pipeline.} Given a textual scene description, (1) the framework first constructs a hierarchical scene proposal mapping functional areas and asset set; (2) semantic constraints are then extracted and refined in a graph via Global Semantic Verification (GSV); finally, (3) a physically valid 3D layout is generated using Global Physical Feasibility Search (GPFS).}
    \vspace{-0.4cm}
    \label{fig:pipeline}
\end{figure*}

\noindent\textbf{Overview.}
As shown in Fig.~\ref{fig:pipeline}, our framework consists of three stages:
\begin{itemize}
    \item[(1)] \textbf{Scene Proposal Construction.} 
    Given the input description, we build a hierarchical scene proposal that separates floor- and furniture-level assets. At each level, a VLM infers functional areas and generates asset descriptions, which are mapped via CLIP-based retrieval to form the final asset set $\mathcal{A}$ with per-asset bounding boxes and category labels.

    \item[(2)] \textbf{Semantic Constraint Modeling and Verification.} 
    The scene proposal is translated into structured inter-asset relational constraints and organized as a scene graph. Global Semantic Verification (GSV) enforces explicit global consistency, eliminating contradictory configurations and yielding a semantically coherent scaffold for layout realization.

    \item[(3)] \textbf{Physically Feasible Layout Realization.} 
    Based on the verified semantic graph, Global Physical Feasibility Search (GPFS) computes a concrete arrangement by combining global exploration with local refinement, ensuring physical validity while preserving semantic relations.
\end{itemize}

Our main contributions lie in the second and third stages of the pipeline, 
namely Global Semantic Verification~(\cref{subsec:gsv}) and 
Gloabl Physical Feasibility Search~(\cref{subsec:hego}). 
We model layout constraints as a graph to capture globally semantically consistent relations, and leverage this relational graph to optimize physically feasible object placements, achieving both search efficiency and solution quality. 


\subsection{Global Semantic Verification}
\label{subsec:gsv}
The Global Semantic Verification (GSV) module aims to enforce global semantic consistency over the relational constraints predicted by the VLM. To achieve this, the verification process is structured into three steps, including scene graph construction and anchor selection, relational consistency verification, and iterative VLM-guided graph refinement. The details of each step are described below.

\subsubsection{Step 1: Scene Graph Construction and Anchor Selection}
\paragraph{\textbf{Scene Graph Construction.}} Building on the hierarchical scene proposal generated by VLM and the asset set $\mathcal{A}$ with their candidate placement regions, the VLM generates a set of implicit pairwise relational constraints in textual form. To obtain a structured representation that enables systematic validation, we formalize these relations as a constraint set and organize them into a directed scene graph. Formally, we define:
\begin{equation}
\mathcal{R} = \{r_{ij} \mid a_i, a_j \in \mathcal{A}\},
\label{eq:constraint_set}
\end{equation}
where each constraint between assets $a_i$ and $a_j$ is defined as $r_{ij} = (a_i, a_j, \tau_{ij},\bm{\psi}_{ij}),$ with $\tau_{ij}$ denoting the constraint type and $\bm{\psi}_{ij}$ denoting the associated parameters (see supplementary material for details).
Based on this formalization, the constraint set $\mathcal{R}$ induces an initial directed graph $G^{(0)}=(V^{(0)},E^{(0)})$, where each node $v_i \in V^{(0)}$ represents an asset $a_i$ with pose $P_i$ and each directed edge $e_{ij} \in E^{(0)}$ encodes a pairwise constraint $r_{ij}$.
The poses $\{P_i\}_{i=1}^N$ are randomly initialized within the room boundary and updated during optimization in Sec.~\ref{subsec:hego}.

\paragraph{\textbf{Anchors Node Selection.}}
Given the initial graph $G^{(0)}$, we partition it to obtain a set of semantically coherent subgraphs $G^{(0)} = \{ G^{(0)}_l\}^N_{l=1}$, where each $G_l^{(0)} = (V_l^{(0)}, E_l^{(0)})$ corresponds to a localized object group. To establish a clear guidance order for conflict detection, each group is anchored by a single central object. Specifically, we define node centrality as in-degree and choose the node with the largest in-degree as the anchor node $v_{l}^{\text{anchor}}$. As a local reference, this anchor has the strongest structural influence on its neighbors and guides subsequent conflict detection. More details on anchor selection are provided in the supplementary material.

\subsubsection{Step 2: Relational Consistency Verification}

Given a refined subgraph $G_l^{(0)}=(V_l^{(0)},E_l^{(0)})$ with its anchor node 
$v_l^{anchor} \in V_l^{(0)}$, we perform subgraph-level validation to detect semantic and structural inconsistencies. 
We introduce a predicate $\Phi$ that evaluates both semantic and structural feasibility. Specifically, $\Phi$ consists of $K$ validation predicate $\{{\Phi_{k}}\}^K_{k=1}$, and the overall validity of the subgraph is defined as:
\begin{equation}
\Phi(G_l) = \bigwedge_{k=1}^{K} \Phi_{k}(G_l),
\end{equation}
where each $\Phi_{k}$ evaluates a semantic criterion on the subgraph $G_l^{(0)}$, $\bigwedge$ denotes logical AND. Consequently, $\Phi(G_l)=1$ if and only if every $\Phi_k$ is satisfied. The $K$ validation predicate $\Phi=\{{\Phi_{k}}\}^K_{k=1}$ cover three aspects: feasibility verification, completeness verification, and semantic consistency verification, as follows.

\paragraph{\textbf{Feasibility Verification.}}
This rule-based module verifies structural and physical validity of the
subgraph using four constraints: cycle consistency ($\Phi_{\text{cycle}}$), distance consistency ($\Phi_{\text{dist}}$), wall compliance ($\Phi_{\text{wall}}$), XY-plane occupancy ($\Phi_{\text{oa}}$) and logic consistency ($\Phi_{\text{logic}}$). Detailed definitions are provided in the supplementary material.

\paragraph{\textbf{Completeness Verification.}}
To ensure relational completeness, we define a completeness predicate $\Phi_{\text{comp}}$. This predicate requires that each node $v_i$ must have at least one positional constraint and one orientation constraint. This is enforced by requiring
$\deg_{\text{pos}}(v_i) \ge 1$ and $\deg_{\text{rot}}(v_i) \ge 1$.

\paragraph{\textbf{Semantic Consistency Verification.}}
We further enforce common-sense semantic priors through pairwise
compatibility checks $\Phi_{\text{sem}}$. 
Violations are detected by:
\begin{equation}
\Psi(G_l) = \prod_{(i,j)\in V_l} (1-\Gamma(e_{ij})),
\end{equation}
where $\Gamma(e_{ij})$ indicates a violated constraint and $\prod$ accumulates all pairwise checks over edges $(i,j)$ in $V_l$.
Detected violations at iteration $t$ are recorded in a conflict log, denoted as $\mathcal{L}^t$, and fed back to the VLM for iterative refinement, producing a conflict-free refined graph $G^*$.

\begin{figure}[!tp]
    \centering
    \includegraphics[width=\textwidth]{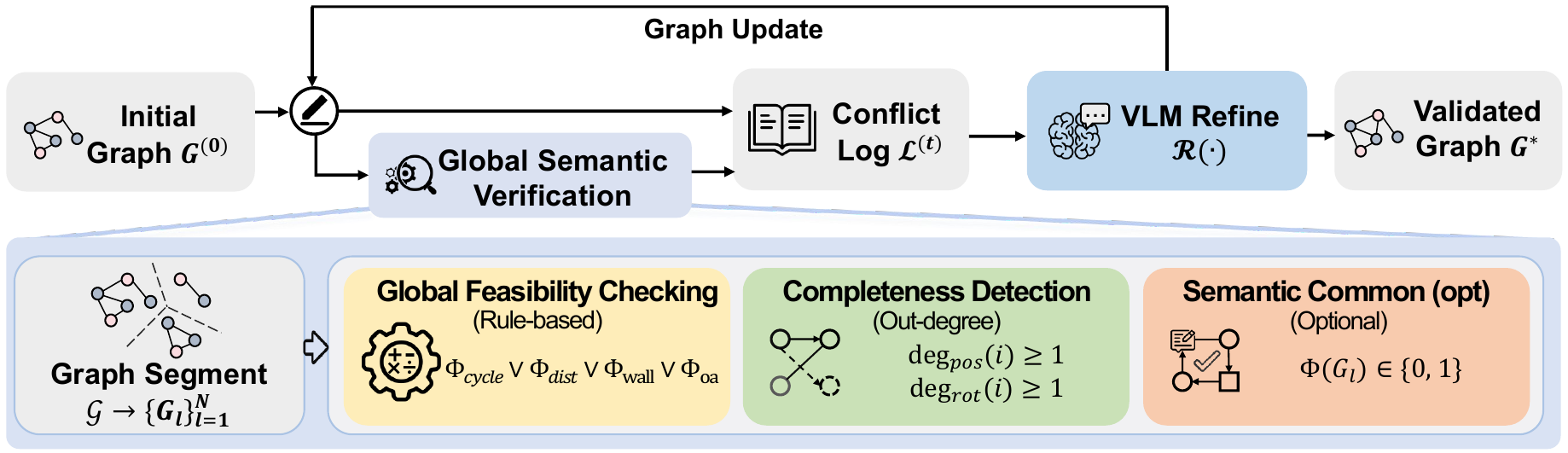}
    \vspace{-0.6cm}
    \caption{\textbf{The Graph-guided Reflection Process.} Through iterative updates to scene-graph edges, the model detects and mitigates physical, semantic, and structural inconsistencies, yielding a coherent and physically plausible layout.}
    \label{fig:graph}
    \vspace{-0.4cm}
\end{figure}

\subsubsection{Step 3: Iterative VLM-Guided Graph Refinement}

We introduce an iterative refinement process that examines the scene graph for physical inconsistencies, semantic contradictions, and missing relational constraints. Detected conflicts are provided to the VLM, which proposes refined edges. This loop repeats until no new conflicts arise, producing a self-consistent constraint graph for subsequent optimization, as illustrated in~\cref{fig:graph}.

Formally, for $E_l^{(t)}$ at iteration $t$, edges identified as conflicting are collected in the set $E_{l, \text{conflict}}^{(t)} \subset E_l^{(t)}$. The VLM then suggests refinements to replace or update these edges, resulting in the next iteration's edge set $E_l^{(t+1)}$. This process continues until no conflicts remain, yielding a consistent graph $G^*$ that serves as stronger guidance for subsequent optimization.


\subsection{Global Physical Feasibility Search}
\label{subsec:hego}

Given the verified relational graph $G^*$ in~\cref{subsec:gsv}, we aim to estimate object poses $\mathcal{P} = \{(x_i, y_i, z_i, \theta_i)\}$ that satisfy the semantic constraints encoded in $G^*$ while remaining physically feasible within the scene.
To achieve robust layout realization in a highly non-convex solution space, the optimization is organized into two steps: global exploration and local refinement. \cref{alg:esgd} provides a detailed description of the full workflow. The following subsections describe the design and workflow of each step in detail.


\begin{algorithm}[!tp]
\caption{\small{Hybrid Evolutionary-Gradient Optimizer}}
\KwIn{
Scene graph $G^{*}$, asset set $\mathcal{A}$, room bound $B$, group set $\mathcal{G}$, anchor nodes $\mathcal{C}$,
parents size $\mu$, offspring size $\lambda$, generations $T$
}
\KwOut{Optimized layout $\mathcal{X}^*$}

\BlankLine
\textbf{Initialization:} Generate $\mu$ parent individuals $\{\mathbf{x}_1, \dots, \mathbf{x}_\mu\}$ via random initialization \\
Evaluate fitness $f_i = \text{Fitness}(\mathbf{x}_i)$ for all parents \\
\BlankLine

\For{$t \leftarrow 1$ \KwTo $T$}{
    \textbf{Recombination and Mutation:}\\
    Select $\lambda$ parents with probability $\propto e^{-\beta f_i}$, swap random asset pairs\\
    \ForEach{offspring $\mathbf{x}$}{
        Apply asset-wise and group-wise gaussian mutation to position/orientation of assets \\
        Apply intra-group attraction toward their center node \\
        Apply pairwise repulsion to avoid collapse \\
        Clamp all positions to room bounds $B$
    }
    \BlankLine
    
    \textbf{Evaluation:}\\
    Combine populations $\mathcal{P} = \{\text{parents} \cup \text{offspring}\}$ \\
    Compute fitness $f(\mathbf{x}_i) = L_{\text{semantic}} + L_{\text{physics}}$  \\
    \BlankLine
    
    \textbf{Selection:}\\
    Retain top-$\mu$ elites by fitness \\
}
\BlankLine

\textbf{Gradient descent:} Perform gradient descent on top-$\mu$ elites for $N$ iterations \\
\textbf{Return} best individual $\mathcal{X}^* = \arg\min_{\mathbf{x}_i} f(\mathbf{x}_i)$
\label{alg:esgd}
\end{algorithm}

\subsubsection{Global Exploration}

Given $G^*$, this step performs a population-based global search to generate diverse candidate layouts that serve as initialization for subsequent local refinement.
To achieve broad coverage of the highly non-convex solution space, we introduce a population-based evolutionary search that maintains multiple layout hypotheses and iteratively improves them through stochastic transformations. Specifically, we design two key operations to guide the exploration process: \emph{\textbf{swap-based crossover}}, which enhances structural diversity by exchanging object poses within a layout, and \emph{\textbf{center-guided mutation}}, which perturbs object placements while preserving group-level spatial coherence. Together, these operations enable effective exploration of feasible layout configurations before local refinement.

\paragraph{\textbf{Swap-Based Crossover.}}
The optimization begins with a randomly initialized population $\{\mathbf{x}_1, \dots, \mathbf{x}_\mu\}$, where $\mu$ denotes the number of parent individuals, each individual $\mathbf{x}_i$ represents a complete scene layout as the set of poses for all $N$ objects:
\begin{equation}
\mathbf{x}_i = \{ P_j^{(i)} \mid j = 1, \dots, N \}.
\end{equation}
At each generation $t$, a set of $\lambda$ offspring is generated through crossover and mutation. 
Traditional crossover can be disruptive in multi-solution settings, as recombination may mix incompatible basins and hinder convergence. We instead use swap-based intra-individual crossover.
Given a selected parent $\mathbf{x}_p$, an intermediate offspring $\mathbf{x}'$ is produced by stochastically swapping the poses of two objects within that parent:
\begin{equation}
\mathbf{x}' = \text{Swap}(\mathbf{x}_p; p_{\text{swap}}),
\end{equation}
where $p_{\text{swap}}$ denotes the probability of performing a swap operation.
This operation is applied primarily at initialization to enhance population diversity and is also activated when fitness stagnates over several generations, allowing the algorithm to escape local minima and prevent early collapse of the solution space. As such, it serves as a crossover mechanism that promotes both initial exploration and robust recovery from suboptimal configurations.

\paragraph{\textbf{Center-Guided Mutation.}}
The intermediate offspring $\mathbf{x}'$ then undergoes Gaussian mutation to produce the final offspring $\mathbf{x}_i$. Specifically, each object pose in $\mathbf{x}'$ is perturbed by zero-mean Gaussian noise, followed by an intra-group attraction:
\begin{equation}
\mathbf{x}_i \leftarrow \mathbf{x}_i + w_{\text{cg}} (\mathbf{c}_g - \mathbf{x}_i),
\end{equation}
where $\mathbf{c}_g$ is the center node of group $g$, and $w_{\text{cg}} \in [0,1]$ controls the attraction toward the group center.
To mitigate solution collapse and maintain population diversity, a pairwise repulsion term is applied as:
\begin{equation}
\mathbf{x}_i \leftarrow \mathbf{x}_i + \eta \sum_{j \neq i} \frac{\mathbf{x}_i - \mathbf{x}_j}{\|\mathbf{x}_i - \mathbf{x}_j\|^2}.
\end{equation}
All positions are subsequently clamped to the room bounds $B$. Center nodes exert stronger influence during optimization, encouraging assets within the same subgraph to form coherent configurations around them.

\subsubsection{Local Refinement}

After the global exploration step identifies a set of promising candidate layouts, 
we perform local refinement to further improve solution quality and ensure precise 
satisfaction of semantic and physical constraints.

\paragraph{\textbf{Elite Selection.}}
After merging the parent and offspring populations, we retain the top-$\mu$ individuals by fitness. 
Each candidate layout $\mathbf{x_i}$ is evaluated by the fitness function:
\begin{equation}
f(\mathbf{x}_i) = L_{\text{semantic}} + L_{\text{physics}}.
\end{equation}
where $L_{\text{semantic}}$ penalizes violations of inter-object semantic relations and $L_{\text{physics}}$ enforces physical feasibility by discouraging collisions and out-of-bound placements. 
These elites provide initialization for the subsequent refinement step.

\paragraph{\textbf{Gradient-Based Refinement.}}
The top-$\mu$ elite individuals are further refined using gradient descent:
\begin{equation}
\mathcal{X}^* = \arg\min_{\mathbf{x}_i} f(\mathbf{x}_i),
\end{equation}
This step refines object poses to better satisfy semantic and physical constraints. 
The resulting configuration $\mathcal{X}^* = \{P_j^* \mid j = 1, \dots, N\}$ forms the final semantically coherent and physically feasible scene layout.

\section{Experiment}

\subsection{Experimental Setup}

\noindent\textbf{Evaluation.}
To validate our method, we follow the benchmark protocol established in LayoutVLM~\cite{Sun25}, conducting experiments across \textbf{11 room types} (a total of \textbf{33 scenes}). We use the same input text instructions as the baseline methods, with assets retrieved via our hierarchical retrieval process, resulting in up to 112 objects per scene.

\noindent\textbf{Evaluation Metrics.}
We evaluate generated layouts in terms of physical plausibility and semantic coherence with respect to the input textual description.
Following LayoutVLM~\cite{Sun25}, we evaluate generated layouts using Collision-Free Score (CF), In-Boundary Score (IB), Positional Coherency (Pos.), Rotational Coherency (Rot.), and Physically-Grounded Semantic Alignment scores(PSA).
Following LayoutVLM~\cite{Sun25}, we employ GPT-4o as a visual-language evaluator, providing both top-down and side-view renderings alongside the input description to simulate human judgments of plausibility and instruction fidelity. All scores are normalized to the range [0, 100], with higher values indicating stronger semantic-physical consistency.

\noindent\textbf{Baselines.}
We compare our method against several recent state-of-the-art approaches for open-vocabulary 3D indoor layout generation, including LayoutGPT~\cite{Feng23}, Holodeck~\cite{Yang24}, I-Design~\cite{Ccelen24}, and LayoutVLM~\cite{Sun25}.

\begin{table}[t]
    \centering
    \resizebox{\linewidth}{!}{%
    \begin{tabular}{l cccc>{\columncolor{lightergray}}c cccc>{\columncolor{lightergray}}c cccc>{\columncolor{lightergray}}c cccc>{\columncolor{lightergray}}c cccc>{\columncolor{lightergray}}c cccc>{\columncolor{lightergray}}c}
    \toprule
    & \multicolumn{5}{c}{\textbf{Bedroom}} & \multicolumn{5}{c}{\textbf{Living Room}} & \multicolumn{5}{c}{\textbf{Dining Room}} & \multicolumn{5}{c}{\textbf{Bookstore}} & \multicolumn{5}{c}{\textbf{Buffet Restaurant}} & \multicolumn{5}{c}{\textbf{Children Room}} \\
    \cmidrule(lr){2-6} \cmidrule(lr){7-11} \cmidrule(lr){12-16} \cmidrule(lr){17-21} \cmidrule(lr){22-26} \cmidrule(lr){27-31}
    \textbf{Methods} & CF & IB & Pos. & Rot. & PSA & CF & IB & Pos. & Rot. & PSA & CF & IB & Pos. & Rot. & PSA & CF & IB & Pos. & Rot. & PSA & CF & IB & Pos. & Rot. & PSA & CF & IB & Pos. & Rot. & PSA \\
    \midrule
    LayoutGPT & 100.0 &  66.7 &  85.7 &  85.9 &  52.2 &  44.4 &  11.1 &  74.7 &  64.4 &  9.6 &  88.9 &  22.2 &  76.0 &  68.9 &  14.8 &  88.9 &  55.6 &  80.9 &  79.4 &  35.9 & 100.0 &  33.3 &  81.2 &  83.3 &  26.9 & 100.0 &   0.0 &  80.9 &  82.6 &   0.0  \\
    Holodeck &  88.9 &  22.2 &  69.3 &  67.9 &  14.1 &  77.8 &   0.0 &  66.3 &  55.6 &   0.0 &  88.9 &   0.0 &  38.0 &  36.6 &   0.0 &  55.6 &   0.0 &  65.7 &  59.0 &   0.0 &  77.8 &  11.1 &  47.7 &  42.4 &   7.4 &  77.8 &  22.2 &  72.7 &  70.0 &  18.7 \\
    I-Design & 100.0 & 77.8 & 72.1 & 65.4 & 51.5 & 33.3 & 11.1 & 62.6 & 46.7 & 0.0 & 88.9 & 66.7 & 76.4 & 66.4 & 34.8 & 66.7 & 11.1 & 68.1 & 69.4 & 5.2 & 100.0 & 55.6 & 63.5 & 57.1 & 35.2 & 77.8 & 55.6 & 78.1 & 75.1 & 34.8 \\
    LayoutVLM & 88.9 & 100.0 &  82.3 &  74.9 & 68.8 &  22.2 &  77.8 &  68.6 &  54.4 &  9.6 & 88.9 & 100.0 &  63.4 &  56.9 &  51.1 & 55.6 & 100.0 &  82.0 &  82.8 &  49.8 & 88.9 &  88.9 &  74.3 &  64.8 & 51.5 & 100.0 & 100.0 &  81.9 &  88.2 &  88.5\\
    \bottomrule
    \textbf{Ours(gpt-4o)} &  100.0 & 100.0  &  83.3 & 80.0 & 73.3 & 100.0 &  100.0 & 95.0 &  62.7 &  80.0  &  100.0 &  100.0   &   95.0   &   88.3   &   80.0   &    100.0   &   100.0   &  89.3  &   90.0   &  \textbf{93.3}   &   100.0   & 66.7  &   81.7   &   70.0    &   66.7   &   100.0   &   100.0  &   88.3   &   88.3   &   80.0   \\
    \textbf{Ours(gpt-4.1)} & 100.0 & 100.0 & 95.0 &  91.7 & \textbf{86.7} & 100.0 & 100.0 &  95.0 &  91.7 &  \textbf{93.3} & 100.0 & 100.0 & 85.0 &  88.3 & \textbf{86.7} & 100.0 & 66.7 &  90.0 & 93.3 & 53.3 & 100.0 &  100.0 &  88.3 & 85.0 & \textbf{86.7} & 100.0 & 100.0 &  90.0 &  88.3 &  \textbf{86.7} \\
    \bottomrule
    \end{tabular}
    }
    \resizebox{\linewidth}{!}{%
    \begin{tabular}{l cccc>{\columncolor{lightergray}}c cccc>{\columncolor{lightergray}}c cccc>{\columncolor{lightergray}}c cccc>{\columncolor{lightergray}}c cccc>{\columncolor{lightergray}}c |cccc>{\columncolor{lightergray}}c}
    \toprule
    & \multicolumn{5}{c}{\textbf{Classroom}} & \multicolumn{5}{c}{\textbf{Computer Room}} & \multicolumn{5}{c}{\textbf{Deli}} & \multicolumn{5}{c}{\textbf{Florist Shop}} & \multicolumn{5}{c|}{\textbf{Game Room}} & \multicolumn{5}{c}{\textbf{Average}} \\
    \cmidrule(lr){2-6} \cmidrule(lr){7-11} \cmidrule(lr){12-16} \cmidrule(lr){17-21} \cmidrule(lr){22-26} \cmidrule(lr){27-31}
    \textbf{Methods} & CF & IB & Pos. & Rot. & PSA & CF & IB & Pos. & Rot. & PSA & CF & IB & Pos. & Rot. & PSA & CF & IB & Pos. & Rot. & PSA & CF & IB & Pos. & Rot. & PSA & CF & IB & Pos. & Rot. & PSA \\
    \midrule
    LayoutGPT & 88.9 &   0.0 &  76.3 &  66.7 &   0.0 & 100.0 &  22.2 &  87.8 &  85.2 &  17.8 &  88.9 &   0.0 &  77.2 &  77.9 &   0.0 &  66.7 &  33.3 &  81.6 &  80.2 &  18.3 &  55.6 &  22.2 &  87.0 &  82.9 &   6.7 & 83.8 &  24.2 &  80.8 & 78.0 &  16.6\\
    Holodeck & 33.3 &   0.0 &  45.2 &  38.6 &   0.0 & 100.0 &   0.0 &  66.1 &  59.7 &   0.0 &  88.9 &  33.3 &  73.9 &  63.7 &  24.4 &  63.9 &   0.0 &  73.2 &  64.7 &   0.0 &  55.6 &  22.2 &  60.7 &  58.0 &   0.0 & 77.8 & 8.1 & 62.8 & 55.6 & 5.6 \\
    I-Design & 55.6 & 11.1 & 50.7 & 47.0 & 0.0 & 88.9 & 22.2 & 74.0 & 70.7 & 8.9 & 88.9 & 22.2 & 67.8 & 65.9 & 10.4 & 77.8 & 0.0 & 75.5 & 68.3 & 0.0 & 66.7 & 44.4 & 62.8 & 58.9 & 17.0 & 76.8 & 34.3 & 68.3 & 62.8 & 18.0 \\
    LayoutVLM & 77.8 & 100.0 &  74.6 &  68.6 &  48.3 & 100.0 &  88.9 &  85.4 &  84.5 & 77.0 & 100.0 &  88.9 &  83.4 &  83.4 & 74.6 & 88.9 & 100.0 &  83.4 &  76.4 & 68.3 & 88.9 & 100.0 &  73.1 & 70.0 &  59.5 & 81.8 & 94.9 & 77.5 & 73.2 & 58.8\\
    \bottomrule
    \textbf{Ours(gpt-4o)} &   100.0   &   100.0    &   81.7   &   88.3   &   73.3   &    100.0  &  100.0  &   70.0   &   73.3    &    66.7    &   66.7   &  100.0   &   81.7   &  85.0    &   \textbf{63.3}   &    100.0   &    100.0   &   85.0   &   91.7   &   \textbf{93.3}   &   100.0   &   100.0   &   86.7   &    87.7   &    93.3   &  97.0  &   \textbf{97.0}  &   85.2   &   82.3   &   78.5   \\
    \textbf{Ours(gpt-4.1)} & 100.0 & 100.0 & 85.0 & 78.3 &  \textbf{91.7} & 100.0 & 100.0 &  96.7 &  95.0 &  \textbf{96.7} & 100.0 & 66.7 & 90.0 &  88.3 & 60.0 & 100.0 & 100.0 & 83.3 &  75.0 &  80.0 & 100.0 & 100.0 &  86.7 &  83.3 &  \textbf{96.7} & \textbf{100.0}  & 93.9 & \textbf{89.5} & \textbf{87.1} & \textbf{83.5} \\
    \bottomrule
    \end{tabular}
    }
    \caption{\textbf{Benchmark Performance.} Performance across all room types under multiple physical and semantic metrics.}
    \label{tab:benchmark_v1}
\end{table}

\begin{figure}[t]
    \centering
    \includegraphics[width=0.97\textwidth]{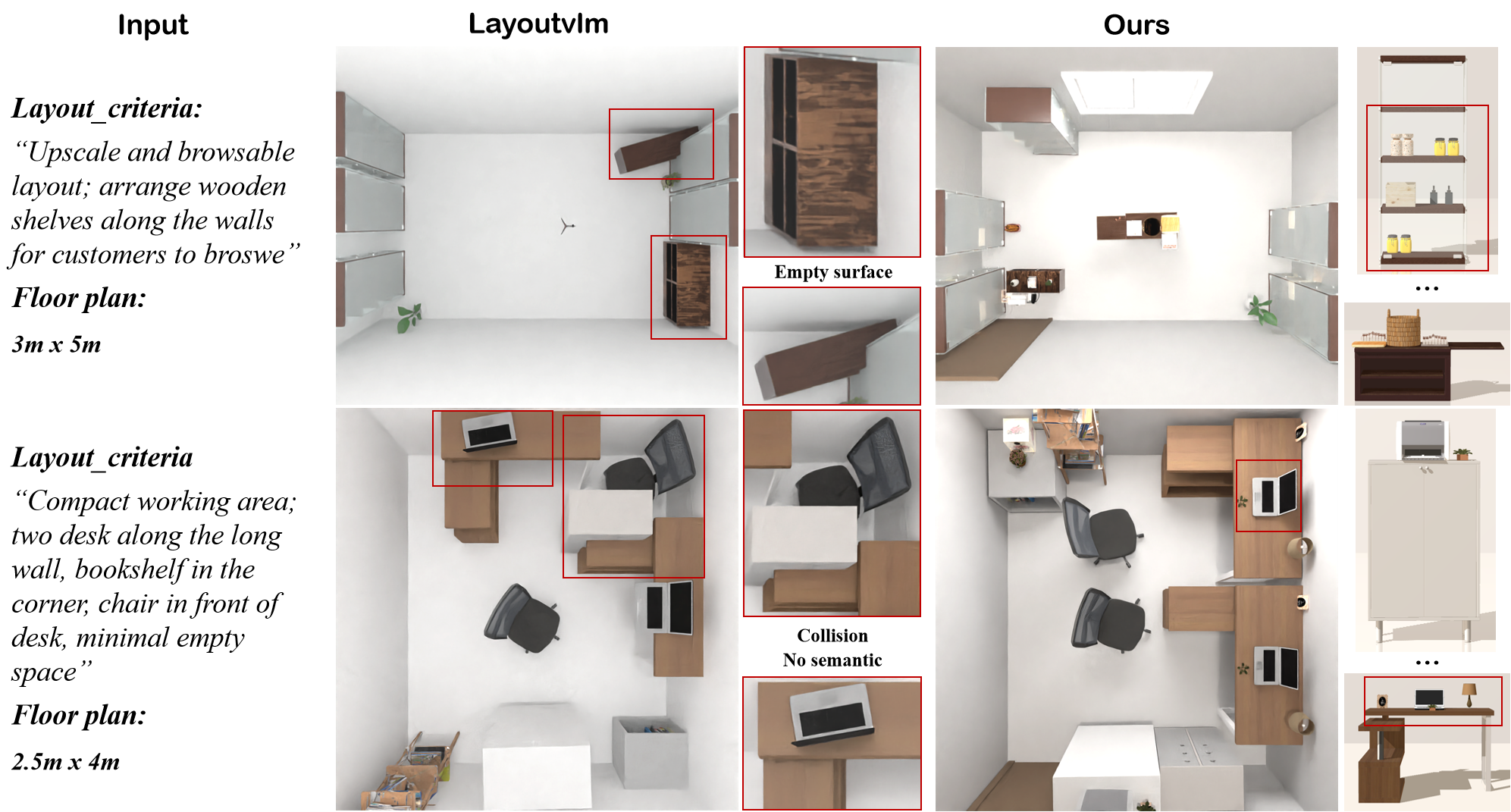}
    \vspace{-0.3cm}
    \caption{\textbf{Qualitative Comparison.} Compared to baseline methods, our approach produces layouts with improved semantic and physical consistency.}
    \label{fig:comparison}
    \vspace{-0.3cm}
\end{figure}
\subsection{Implementation Details}
We employ GPT-4o and GPT-4.1 as the VLM backbone for semantic reasoning and constraint generation. For asset retrieval, we use the 3D-FUTURE\cite{Fu21b} and Objaverse\cite{Deitke23} datasets for floor-level objects, and the HSSD-200\cite{Khanna24} dataset for furniture-level object retrieval.
In the evolutionary stage, each scene region is initialized with random object positions. 
The population size is set to 50, with 5 parent individuals retained per generation. 
Regions containing fewer than 10 objects are optimized for 30 generations, while larger regions run for 50 generations. 
The top 5 layouts with the lowest objective energy are selected as initial states for gradient-based refinement, followed by 200 steps of gradient descent. 
The final layout is the configuration with minimal objective energy.

\subsection{Benchmark Performance}
\cref{tab:benchmark_v1} reports comparisons against LayoutGPT, Holodeck, I-Design, and LayoutVLM. Our method achieves higher scores across all metrics (CF, IB, Pos., Rot., PSA) and all room types, indicating stronger semantic coherence and higher physical feasibility across diverse layout scenarios.

\begin{figure}[!tp]
    \centering
    \includegraphics[width=.9\textwidth]{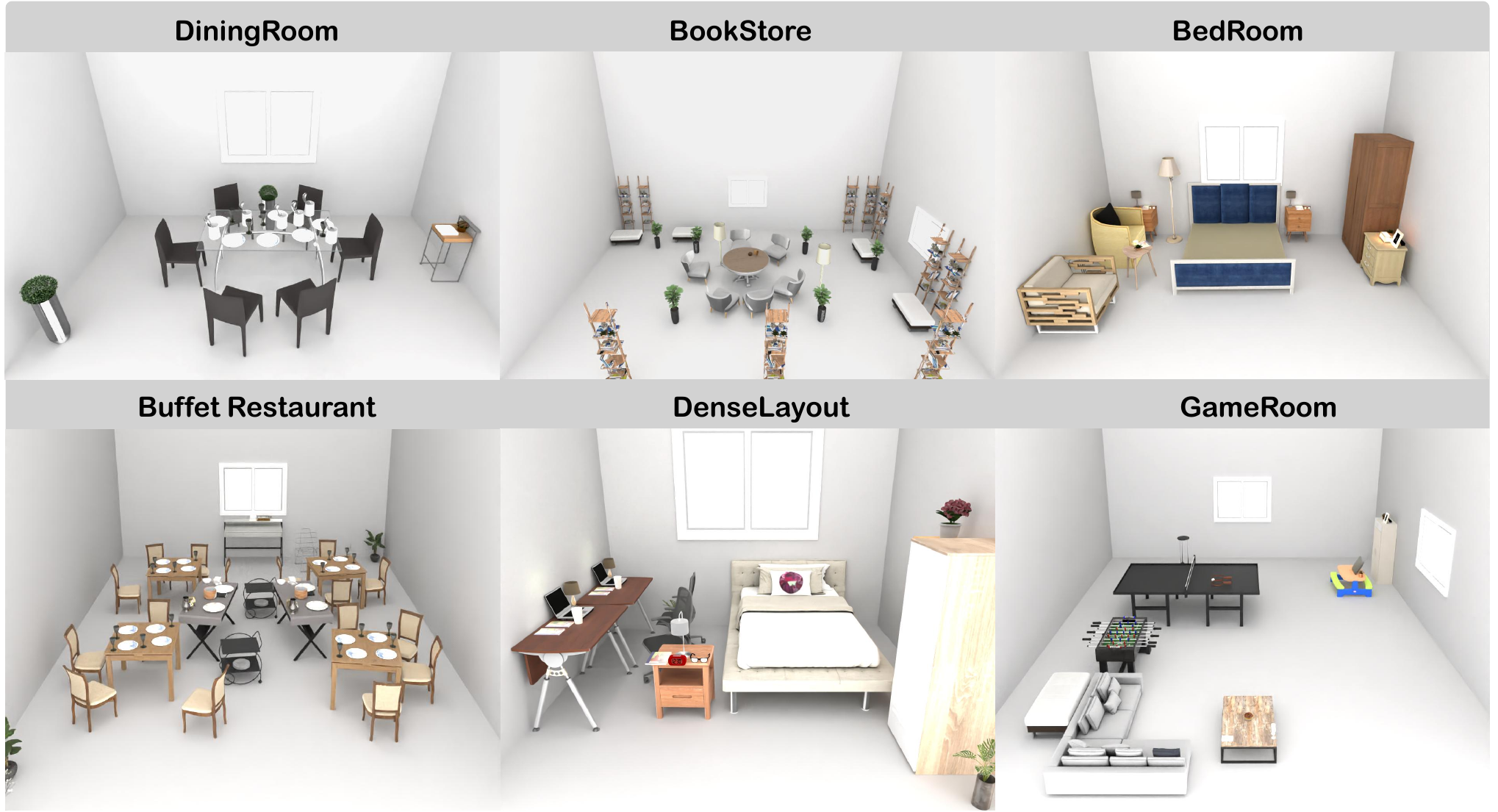}
    \vspace{-0.1cm}
        \caption{\textbf{Results.} the results demonstrate that the proposed method can generate dense and collision-free indoor layouts}
    \label{fig:results}
    \vspace{-0.5em}
\end{figure}

The high CF and IB scores mainly arise from the Global Physical Feasibility Search (GPFS) module, which combines global exploration with local refinement to prevent the optimizer from being trapped in local minima, 
thereby reducing object collisions and placements outside the room. Under this hybrid search strategy, even highly cluttered or entangled initial layouts can be optimized into physically valid configurations, as demonstrated in buffer\_restaurant and dense\_layout of~\cref{fig:results}, where densely packed objects are effectively rearranged without collisions, highlighting GPFS’s robustness to challenging initializations.

Improvements in Pos. and Rot. further indicate stronger semantic alignment with textual instructions. This is primarily enabled by our Global Semantic Verification (GSV) module, which explicitly enforces hierarchical and relational semantics at the graph level, preventing inconsistent or implausible object–object relationships early in the process and producing coherent spatial relations. Combined with effective optimization, this mechanism ensures that spatial directives are faithfully realized in the final configuration and promotes functionally coherent arrangements.

\begin{figure}[!t]
    \centering
    \vspace{-0.1cm}
    \includegraphics[width=.9\textwidth]{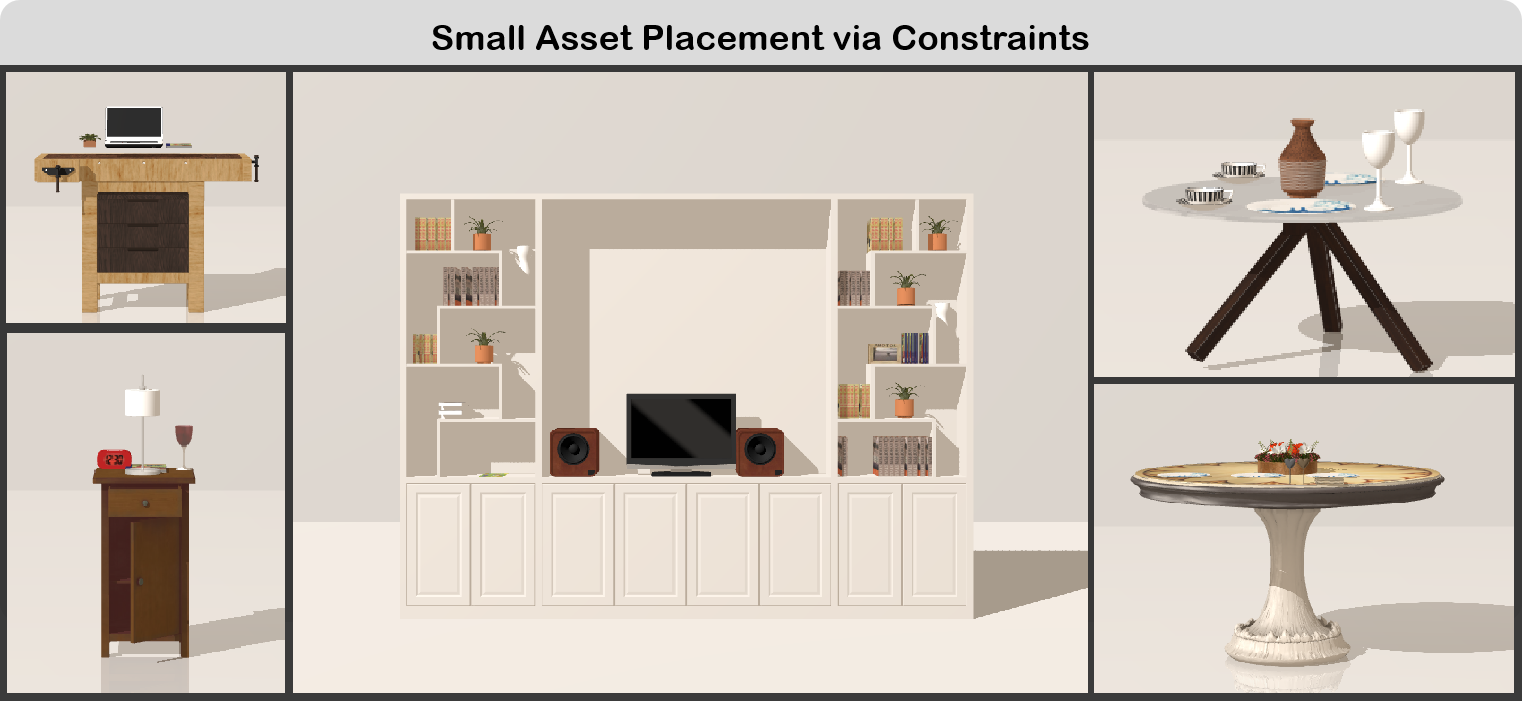}
    \caption{\textbf{Furniture surface placement.} Small asset placement guided by spatial constraints.}
    \label{fig:surface_result}
    \vspace{-0.4cm}
\end{figure}

The gains in PSA reflect the unified effect of semantic validation and physically grounded optimization. The GSV module guarantees that semantic constraints are globally consistent and feasible, while GPFS ensures that these constraints remain physically realizable throughout optimization. As a result, the generated layouts are simultaneously semantically faithful, functionally plausible, and physically coherent, leading to the highest overall PSA performance. This behavior is consistently observed at both the floor-object and furniture-object levels (see~\cref{fig:results,fig:surface_result}), where our framework produces correct relational structures with negligible collisions across different object granularities.

Nevertheless, both our GPT-4o and GPT-4.1 variants still exhibit a few failure cases, particularly in the Bookstore and Deli scenes. These scenes are challenging for physical optimization for two main reasons. First, they often involve crowded layouts with relatively large assets and dense object placements on furniture surfaces, making fine-grained collision and boundary optimization difficult. Second, many objects are wall-aligned (e.g., bookshelves) and constrained by near-wall relations, resulting in a highly constrained optimization landscape.

\vspace{-1.6em}
\begin{table}[h!]
    \centering\small
    \resizebox{\textwidth}{!}{%
    \setlength{\tabcolsep}{5pt} 
    \renewcommand{\arraystretch}{1.05}
    \begin{tabular}{l lccccc}
    \hline
    Module & Method & \multicolumn{2}{c}{Physics} & \multicolumn{2}{c}{Semantics} & Overall score \\
    \cline{3-4} \cline{5-6} \cline{7-7}
    & & CF & IB & Pos. & Rot. & PSA\\
    \hline
    \multirow{4}{*}{GSV} 
    & Ours & \textbf{100.0} & \textbf{100.0} & \textbf{91.3} & 87.7 & \textbf{86.7} \\
    & w/o Feasibility & 93.3 & \textbf{100.0} & 88.7 & \textbf{90.3} & 82.7 \\
    & w/o Completeness & 93.3 & 93.3 & 89.0 & 86.7 & 74.0 \\
    & w/o Semantic Consistency & \textbf{100.0} & \textbf{100.0} & 90.0 & 89.1 & 85.3 \\
    \midrule
    \multirow{2}{*}{GPFS} 
    & w/o GD & 13.3 & \textbf{0.0} & 79.7 & 79.6 & 0.0 \\
    & w/o EA & 73.3 & \textbf{100.0} & 86.0 & 80.5 & 61.3 \\
    \hline
    \end{tabular}
    }
    \vspace{5pt}
    \caption{\textbf{Ablation study.} Quantitative results grouped by global semantic verification (GSV) and global physical feasibility search (GPFS) modules.}
    \label{tab:ablation}
\end{table}
\vspace{-2.0em}

\subsection{Ablation Study}
We conduct ablation studies on a representative subset of scenes, focusing on those that are prone to conflicts or local optima. Specifically, this subset covers five scene types: Bedroom, FloristShop, BuffetRestaurant, LivingRoom, and DiningRoom. These scenes span diverse spatial characteristics and relational complexities, including constrained layouts, functionally coupled object groups, and relatively high interaction density. Such diversity ensures that the effects of each component can be clearly and reliably assessed, while avoiding dilution from trivially solvable or conflict-free cases.
As shown in~\cref{tab:ablation}, each component distinctly improves semantic quality and physical validity.

\begin{figure}[t]
    \centering
    \vspace{-0.3cm}
    \includegraphics[width=.9\textwidth]{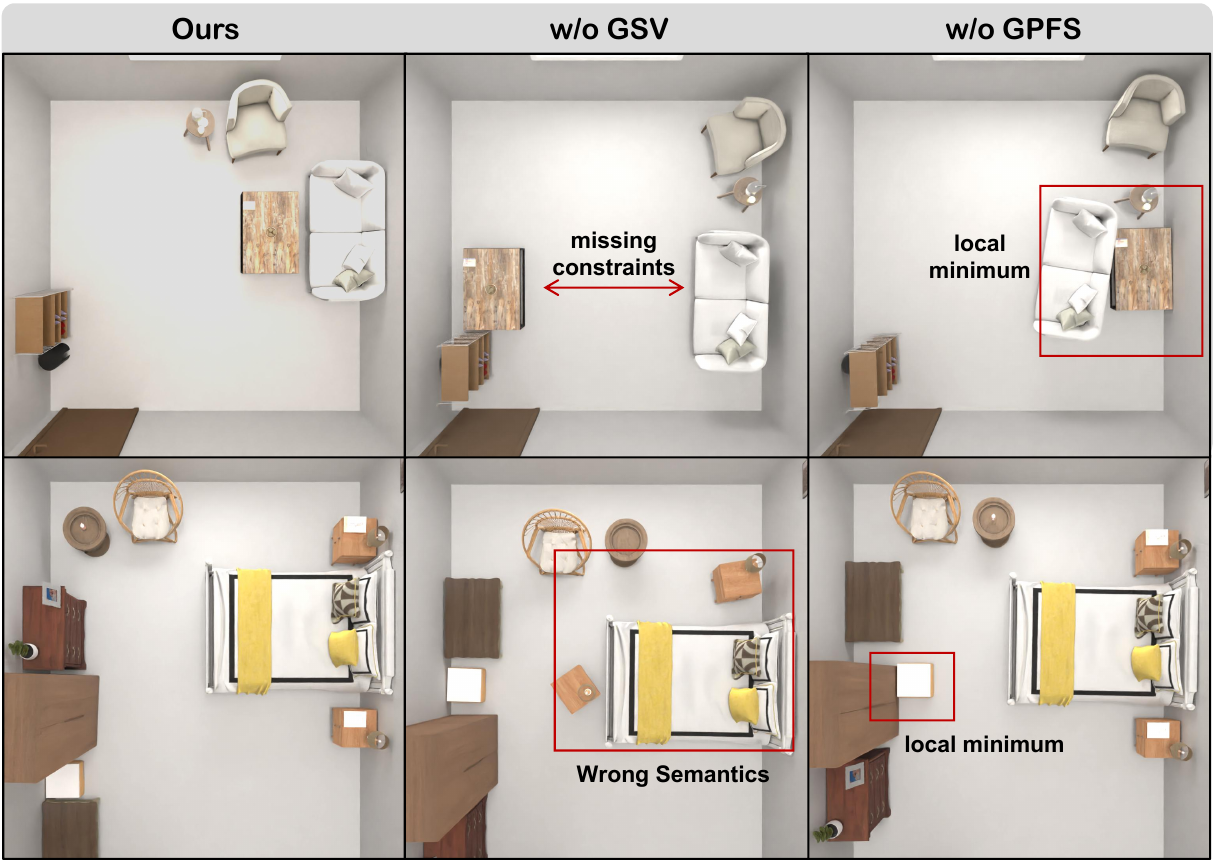}
    \vspace{-0.2cm}
    \caption{\textbf{Ablation study.} Comparison of layout results with or without GSV and GPFS module, indicating that GSV ensures semantic coherence and GPFS promotes globally feasible layouts.}
    \vspace{-1em}
    \label{fig:ablation}
    \vspace{-1em}
\end{figure}

\noindent\textbf{Effect of Global Semantic Verification.}
Removing aspects of the graph module leads to consistent performance degradation. Without feasibility filtering, CF drops from 100.0 to 93.3 and PSA decreases to 82.7, showing that physically infeasible constraints cause unstable interactions during optimization. Removing constraint completeness causes a larger PSA drop (86.7 $\rightarrow$ 74.0), indicating that incomplete relational specifications produce semantically under-constrained layouts. Disabling semantic consistency checking also reduces PSA (to 85.3), highlighting the importance of globally coherent relational structures. These trends are visually confirmed in~\cref{fig:ablation}, where missing or inconsistent constraints cause misaligned object groupings and incorrect relational configurations.

\noindent\textbf{Effect of Global Physical Feasibility Search.}
The impact of the hybrid evolutionary–gradient optimization is even more pronounced. Removing gradient refinement (w/o GD) causes CF to collapse to 13.3, IB to 0.0 and PSA to 0.0, indicating that evolutionary search alone cannot ensure physically feasible convergence. Conversely, removing evolutionary search (w/o EA) reduces CF to 73.3 and PSA to 61.3, demonstrating that gradient descent without global exploration frequently becomes trapped in poor local minima. As illustrated in~\cref{fig:ablation}, the pure GD variant produces locally entangled layouts, while the absence of GD leaves unresolved collisions.
Overall, the ablation results quantitatively and qualitatively confirm that graph-based constraint reasoning ensures semantic completeness and consistency, while the hybrid evolutionary-gradient strategy is critical for achieving physically valid and globally coherent layouts.

\section{Conclusion and Limitation}

We revisit VLM-based 3D indoor layout generation from a global reasoning perspective and identify two key limitations: the lack of explicit scene-level semantic verification and the reliance on purely local optimization in a highly non-convex feasible space. To address these issues, we disentangle semantic coherence from physical feasibility through two dedicated mechanisms. Global Semantic Verification (GSV) enforces explicit structural consistency over scene graphs, while Global Physical Feasibility Search (GPFS) combines evolutionary exploration with gradient refinement to enable robust optimization beyond local minima. 

Extensive experiments demonstrate state-of-the-art performance in both semantic consistency and physical plausibility. Ablation studies further confirm that explicit graph validation is critical for global coherence, and hybrid global
local search is essential for physically valid layout realization. Together, our framework shifts layout generation from local relational fitting to globally consistent and physically grounded reasoning, advancing the reliability of open-vocabulary 3D indoor layout generation.

Despite the promising results, several limitations remain. First, our pipeline still relies on GPT APIs, whose inherent stochasticity affects object category selection, object quantity, and the subsequent evaluation process, leading to variations across different runs. 
Second, the quality of current 3D asset collections is highly inconsistent. Some assets are oversized, while others may fail to render correctly, which can significantly affect the generated layouts. 
Although these problematic assets are relatively easy to identify, incorporating a lightweight asset pre-check or filtering stage would substantially improve the robustness and success rate of large-scale scene generation. 
Reducing pipeline stochasticity and constructing higher-quality 3D asset collections therefore remain important directions for future work.

\section{Acknowledgement}
This research was supported in part by the National Natural Science Foundation of China (62525115, U2570208) and by the Fundamental Research Funds for the Central Universities (G2026KY05065).

%
%
\bibliographystyle{splncs04}
\bibliography{bibtex/string,bibtex/short,bibtex/vision,bibtex/graphic}

@string{AI     = "AI" }

@STRING{ARXIV  = "arXiv Preprint"}

@string{SIGGRAPH = "SIGGRAPH" }

@string{TOG    = "TOG" }

@STRING{AI       = "Artificial Intelligence"}

@STRING{ACM      = "Association for Computing Machinery"}

@STRING{ARXIV    = "arXiv Preprint"}

@STRING{EMNLP     = "Conference on Empirical Methods in Natural Language Processing"}

@STRING{SIGGRAPH = "ACM SIGGRAPH"}

@STRING{TOG      = "ACM Transactions on Graphics"}

@article{Aguina24,
  title={Open-universe indoor scene generation using llm program synthesis and uncurated object databases},
  author={Aguina-Kang, Rio and Gumin, Maxim and Han, Do Heon and Morris, Stewart and Yoo, Seung Jean and Ganeshan, Aditya and Jones, R Kenny and Wei, Qiuhong Anna and Fu, Kailiang and Ritchie, Daniel},
  journal={arXiv preprint arXiv:2403.09675},
  year={2024}
}

@inproceedings{Armeni19,
  title={3d scene graph: A structure for unified semantics, 3d space, and camera},
  author={Armeni, Iro and He, Zhi-Yang and Gwak, JunYoung and Zamir, Amir R and Fischer, Martin and Malik, Jitendra and Savarese, Silvio},
  booktitle={Proceedings of the IEEE/CVF international conference on computer vision},
  pages={5664--5673},
  year={2019}
}

@book{Boyd04,
  title={Convex optimization},
  author={Boyd, Stephen and Vandenberghe, Lieven},
  year={2004},
  publisher={Cambridge university press}
}

@inproceedings{Ccelen24,
  title={I-design: Personalized llm interior designer},
  author={{\c{C}}elen, Ata and Han, Guo and Schindler, Konrad and Van Gool, Luc and Armeni, Iro and Obukhov, Anton and Wang, Xi},
  booktitle={European Conference on Computer Vision},
  pages={217--234},
  year={2024},
  organization={Springer}
}

@inproceedings{Chang14,
  title={Learning spatial knowledge for text to 3D scene generation},
  author={Chang, Angel and Savva, Manolis and Manning, Christopher D},
  booktitle={Proceedings of the 2014 conference on empirical methods in natural language processing (EMNLP)},
  pages={2028--2038},
  year={2014}
}

@article{Chang17,
  title={SceneSeer: 3D scene design with natural language},
  author={Chang, Angel X and Eric, Mihail and Savva, Manolis and Manning, Christopher D},
  journal={arXiv preprint arXiv:1703.00050},
  year={2017}
}

@inproceedings{Deitke23,
  title={Objaverse: A universe of annotated 3d objects},
  author={Deitke, Matt and Schwenk, Dustin and Salvador, Jordi and Weihs, Luca and Michel, Oscar and VanderBilt, Eli and Schmidt, Ludwig and Ehsani, Kiana and Kembhavi, Aniruddha and Farhadi, Ali},
  booktitle={Proceedings of the IEEE/CVF Conference on Computer Vision and Pattern Recognition},
  pages={13142--13153},
  year={2023}
}

@inproceedings{Deng25,
  title={Global-local tree search in vlms for 3d indoor scene generation},
  author={Deng, Wei and Qi, Mengshi and Ma, Huadong},
  booktitle={Proceedings of the Computer Vision and Pattern Recognition Conference},
  pages={8975--8984},
  year={2025}
}

@inproceedings{Dhamo21,
  title={Graph-to-3d: End-to-end generation and manipulation of 3d scenes using scene graphs},
  author={Dhamo, Helisa and Manhardt, Fabian and Navab, Nassir and Tombari, Federico},
  booktitle={Proceedings of the IEEE/CVF International Conference on Computer Vision},
  pages={16352--16361},
  year={2021}
}

@article{Feng23,
  title={Layoutgpt: Compositional visual planning and generation with large language models},
  author={Feng, Weixi and Zhu, Wanrong and Fu, Tsu-jui and Jampani, Varun and Akula, Arjun and He, Xuehai and Basu, Sugato and Wang, Xin Eric and Wang, William Yang},
  journal={Advances in Neural Information Processing Systems},
  volume={36},
  pages={18225--18250},
  year={2023}
}

@incollection{Fisher11,
  title={Characterizing structural relationships in scenes using graph kernels},
  author={Fisher, Matthew and Savva, Manolis and Hanrahan, Pat},
  booktitle={ACM SIGGRAPH 2011 papers},
  pages={1--12},
  year={2011},
  publisher={ACM}
}

@article{Fisher12,
  title={Example-based synthesis of 3D object arrangements},
  author={Fisher, Matthew and Ritchie, Daniel and Savva, Manolis and Funkhouser, Thomas and Hanrahan, Pat},
  journal={ACM Transactions on Graphics (TOG)},
  volume={31},
  number={6},
  pages={1--11},
  year={2012},
  publisher={ACM New York, NY, USA}
}

@article{Fu17,
  title={Adaptive synthesis of indoor scenes via activity-associated object relation graphs},
  author={Fu, Qiang and Chen, Xiaowu and Wang, Xiaotian and Wen, Sijia and Zhou, Bin and Fu, Hongbo},
  journal={ACM Transactions on Graphics (TOG)},
  volume={36},
  number={6},
  pages={1--13},
  year={2017},
  publisher={ACM New York, NY, USA}
}

@inproceedings{Fu21a,
  title={3d-front: 3d furnished rooms with layouts and semantics},
  author={Fu, Huan and Cai, Bowen and Gao, Lin and Zhang, Ling-Xiao and Wang, Jiaming and Li, Cao and Zeng, Qixun and Sun, Chengyue and Jia, Rongfei and Zhao, Binqiang and others},
  booktitle={Proceedings of the IEEE/CVF International Conference on Computer Vision},
  pages={10933--10942},
  year={2021}
}

@article{Fu21b,
  title={3D-FUTURE: 3D furniture shape with texture},
  author={Fu, Hongbo and Jia, Ruizhi and Gao, Ling and et al.},
  journal={International Journal of Computer Vision},
  volume={129},
  number={12},
  pages={3313--3337},
  year={2021},
  publisher={Springer}
}

@inproceedings{Fu24,
  title={Anyhome: Open-vocabulary generation of structured and textured 3d homes},
  author={Fu, Rao and Wen, Zehao and Liu, Zichen and Sridhar, Srinath},
  booktitle={European Conference on Computer Vision},
  pages={52--70},
  year={2024},
  organization={Springer}
}

@article{Hu24,
  title={Mixed diffusion for 3d indoor scene synthesis},
  author={Hu, Siyi and Arroyo, Diego Martin and Debats, Stephanie and Manhardt, Fabian and Carlone, Luca and Tombari, Federico},
  journal={arXiv preprint arXiv:2405.21066},
  year={2024}
}

@inproceedings{Johnson18,
  title={Image generation from scene graphs},
  author={Johnson, Justin and Gupta, Agrim and Fei-Fei, Li},
  booktitle={Proceedings of the IEEE conference on computer vision and pattern recognition},
  pages={1219--1228},
  year={2018}
}

@inproceedings{Khanna24,
  title={Habitat synthetic scenes dataset (hssd-200): An analysis of 3d scene scale and realism tradeoffs for objectgoal navigation},
  author={Khanna, Mukul and Mao, Yongsen and Jiang, Hanxiao and Haresh, Sanjay and Shacklett, Brennan and Batra, Dhruv and Clegg, Alexander and Undersander, Eric and Chang, Angel X and Savva, Manolis},
  booktitle={Proceedings of the IEEE/CVF Conference on Computer Vision and Pattern Recognition},
  pages={16384--16393},
  year={2024}
}

@article{Kingma13,
    title={Auto-encoding variational bayes},
    author={Kingma, Diederik P. and Welling, Max},
    journal={arXiv preprint arXiv:1312.6114},
    year={2013}
}

@article{Li19b,
  title={Grains: Generative recursive autoencoders for indoor scenes},
  author={Li, Manyi and Patil, Akshay Gadi and Xu, Kai and Chaudhuri, Siddhartha and Khan, Owais and Shamir, Ariel and Tu, Changhe and Chen, Baoquan and Cohen-Or, Daniel and Zhang, Hao},
  journal={ACM Transactions on Graphics (TOG)},
  volume={38},
  number={2},
  pages={1--16},
  year={2019},
  publisher={ACM New York, NY, USA}
}

@article{Lin24,
  title={Instructscene: Instruction-driven 3d indoor scene synthesis with semantic graph prior},
  author={Lin, Chenguo and Mu, Yadong},
  journal={arXiv preprint arXiv:2402.04717},
  year={2024}
}

@article{Ling25,
  title={Scenethesis: A language and vision agentic framework for 3d scene generation},
  author={Ling, Lu and Lin, Chen-Hsuan and Lin, Tsung-Yi and Ding, Yifan and Zeng, Yu and Sheng, Yichen and Ge, Yunhao and Liu, Ming-Yu and Bera, Aniket and Li, Zhaoshuo},
  journal={arXiv preprint arXiv:2505.02836},
  year={2025}
}

@inproceedings{Littlefair25,
  title={FlairGPT: Repurposing LLMs for interior designs},
  author={Littlefair, Gabrielle and Dutt, Niladri Shekhar and Mitra, Niloy J},
  booktitle={Computer Graphics Forum},
  pages={e70036},
  year={2025},
  organization={Wiley Online Library}
}

@article{Liu14,
  title={Creating consistent scene graphs using a probabilistic grammar},
  author={Liu, Tianqiang and Chaudhuri, Siddhartha and Kim, Vladimir G and Huang, Qixing and Mitra, Niloy J and Funkhouser, Thomas},
  journal={ACM Transactions on Graphics (TOG)},
  volume={33},
  number={6},
  pages={1--12},
  year={2014},
  publisher={ACM New York, NY, USA}
}

@inproceedings{Luo20,
  title={End-to-end optimization of scene layout},
  author={Luo, Andrew and Zhang, Zhoutong and Wu, Jiajun and Tenenbaum, Joshua B},
  booktitle={Proceedings of the IEEE/CVF Conference on Computer Vision and Pattern Recognition},
  pages={3754--3763},
  year={2020}
}

@article{Merrell11,
  title={Interactive furniture layout using interior design guidelines},
  author={Merrell, Paul and Schkufza, Eric and Li, Zeyang and Agrawala, Maneesh and Koltun, Vladlen},
  journal={ACM transactions on graphics (TOG)},
  volume={30},
  number={4},
  pages={1--10},
  year={2011},
  publisher={ACM New York, NY, USA}
}

@inproceedings{Purkait20,
  title={Sg-vae: Scene grammar variational autoencoder to generate new indoor scenes},
  author={Purkait, Pulak and Zach, Christopher and Reid, Ian},
  booktitle={European Conference on Computer Vision},
  pages={155--171},
  year={2020},
  organization={Springer}
}

@article{Pun25,
  title={HSM: Hierarchical Scene Motifs for Multi-Scale Indoor Scene Generation},
  author={Pun, Hou In Derek and Tam, Hou In Ivan and Wang, Austin T and Huo, Xiaoliang and Chang, Angel X and Savva, Manolis},
  journal={arXiv preprint arXiv:2503.16848},
  year={2025}
}

@article{Sohn15,
    title={Learning structured output representation using deep conditional generative models},
    author={Sohn, Kihyuk and Lee, Honglak and Yan, Xinchen},
    journal={Advances in Neural Information Processing Systems},
    volume={28},
    year={2015}
}

@article{Su25,
  title={Chord: Generation of collision-free, house-scale, and organized digital twins for 3d indoor scenes with controllable floor plans and optimal layouts},
  author={Su, Chong and Fu, Yingbin and Hu, Zheyuan and Yang, Jing and Hanji, Param and Wang, Shaojun and Zhao, Xuan and {\"O}ztireli, Cengiz and Zhong, Fangcheng},
  journal={arXiv preprint arXiv:2503.11958},
  year={2025}
}

@inproceedings{Sun24,
  title={Forest2seq: Revitalizing order prior for sequential indoor scene synthesis},
  author={Sun, Qi and Zhou, Hang and Zhou, Wengang and Li, Li and Li, Houqiang},
  booktitle={European Conference on Computer Vision},
  pages={251--268},
  year={2024},
  organization={Springer}
}

@inproceedings{Sun25,
  title={Layoutvlm: Differentiable optimization of 3d layout via vision-language models},
  author={Sun, Fan-Yun and Liu, Weiyu and Gu, Siyi and Lim, Dylan and Bhat, Goutam and Tombari, Federico and Li, Manling and Haber, Nick and Wu, Jiajun},
  booktitle={Proceedings of the Computer Vision and Pattern Recognition Conference},
  pages={29469--29478},
  year={2025}
}

@inproceedings{Tang24,
  title={Diffuscene: Denoising diffusion models for generative indoor scene synthesis},
  author={Tang, Jiapeng and Nie, Yinyu and Markhasin, Lev and Dai, Angela and Thies, Justus and Nie{\ss}ner, Matthias},
  booktitle={Proceedings of the IEEE/CVF conference on computer vision and pattern recognition},
  pages={20507--20518},
  year={2024}
}

@inproceedings{Wald20,
  title={Learning 3d semantic scene graphs from 3d indoor reconstructions},
  author={Wald, Johanna and Dhamo, Helisa and Navab, Nassir and Tombari, Federico},
  booktitle={Proceedings of the IEEE/CVF Conference on Computer Vision and Pattern Recognition},
  pages={3961--3970},
  year={2020}
}

@inproceedings{Wang21c,
  title={Sceneformer: Indoor scene generation with transformers},
  author={Wang, Xinpeng and Yeshwanth, Chandan and Nie{\ss}ner, Matthias},
  booktitle={2021 International Conference on 3D Vision (3DV)},
  pages={106--115},
  year={2021},
  organization={IEEE}
}

@inproceedings{Wu24,
  title={External knowledge enhanced 3d scene generation from sketch},
  author={Wu, Zijie and Feng, Mingtao and Wang, Yaonan and Xie, He and Dong, Weisheng and Miao, Bo and Mian, Ajmal},
  booktitle={European Conference on Computer Vision},
  pages={286--304},
  year={2024},
  organization={Springer}
}

@inproceedings{Yang24,
  title={Holodeck: Language guided generation of 3d embodied ai environments},
  author={Yang, Yue and Sun, Fan-Yun and Weihs, Luca and VanderBilt, Eli and Herrasti, Alvaro and Han, Winson and Wu, Jiajun and Haber, Nick and Krishna, Ranjay and Liu, Lingjie and others},
  booktitle={Proceedings of the IEEE/CVF Conference on Computer Vision and Pattern Recognition},
  pages={16227--16237},
  year={2024}
}

@inproceedings{Yang24a,
  title={Physcene: Physically interactable 3d scene synthesis for embodied ai},
  author={Yang, Yandan and Jia, Baoxiong and Zhi, Peiyuan and Huang, Siyuan},
  booktitle={Proceedings of the IEEE/CVF Conference on Computer Vision and Pattern Recognition},
  pages={16262--16272},
  year={2024}
}

@article{Yeh12,
  title={Synthesizing open worlds with constraints using locally annealed reversible jump mcmc},
  author={Yeh, Yi-Ting and Yang, Lingfeng and Watson, Matthew and Goodman, Noah D and Hanrahan, Pat},
  journal={ACM Transactions on Graphics (TOG)},
  volume={31},
  number={4},
  pages={1--11},
  year={2012},
  publisher={ACM New York, NY, USA}
}

@article{Zhao11,
  title={Image parsing with stochastic scene grammar},
  author={Zhao, Yibiao and Zhu, Song-Chun},
  journal={Advances in Neural Information Processing Systems},
  volume={24},
  year={2011}
}

@inproceedings{Zhou19b,
  title={Scenegraphnet: Neural message passing for 3d indoor scene augmentation},
  author={Zhou, Yang and While, Zachary and Kalogerakis, Evangelos},
  booktitle={Proceedings of the IEEE/CVF International Conference on Computer Vision},
  pages={7384--7392},
  year={2019}
}

\clearpage
\appendix
\begin{center}
{\Large\bfseries Global Graph-Validated Optimization for}

{\Large\bfseries VLM-based 3D Indoor Scene Generation}

{\Large\bfseries -- Supplementary Materials --}


\end{center}

\setcounter{page}{1}
\setcounter{figure}{0}
\setcounter{table}{0}
\thispagestyle{empty}

\startcontents[supp]
\begingroup
\setcounter{tocdepth}{2}
\makeatletter
\renewcommand{\contentsname}{}
\makeatother
\printcontents[supp]{}{1}{}
\endgroup

\section{Additional results}

\begin{table}[h]
\centering
\vspace{-5pt}
\setlength{\tabcolsep}{1.4pt}
\resizebox{\linewidth}{!}{%
\begin{tabular}{l|cccc|cc|cccc|c}
\toprule
& \multicolumn{4}{c|}{SceneEval Fidelity (\%)} 
& \multicolumn{6}{c|}{SceneEval Plausibility (\%)} 
& \multirow{2}{*}{Time} \\
\cmidrule(lr){2-5} \cmidrule(lr){6-11}
Method 
& $\uparrow$ CNT 
& $\uparrow$ ATR 
& $\uparrow$ OOR
& $\uparrow$ OAR 
& $\downarrow$ COL$_{ob}$ 
& $\downarrow$ COL$_{sc}$ 
& $\uparrow$ SUP
& $\uparrow$ NAV
& $\uparrow$ ACC 
& $\downarrow$ OOB
&  \\
\midrule
LayoutVLM 
& 41.19 & 22.26 & 8.60 & 23.29 
& 36.09 & 69.00 
& 67.96 & 98.75 & 85.91 & 4.14 
& 247.4\,s \\

Holodeck 
& 43.98 & 40.10 & 19.87 & 48.26 
& 17.85 & 72.00 
& 61.38 & \textbf{99.51} & \textbf{90.23} & \textbf{1.12}
& 133.5 s \\

Ours 
& \textbf{52.65} & \textbf{51.54} & \textbf{25.61} & \textbf{49.41} 
& \textbf{7.15} & \textbf{36.00} 
& \textbf{84.35} & 93.25 & 65.73 & 2.09 
& 302.3\,s \\
\bottomrule
\end{tabular}
}
\vspace{2pt}

\caption{Comparison on the larger-scale SceneEval-100 containing 100 scenes with diverse scene types and complexity levels. Runtime is reported using the same generation mode described in the original LayoutVLM paper.}
\label{tab:sceneeval}
\vspace{-5pt}
\end{table}
In addition to the LayoutVLM benchmark, we evaluate on the larger SceneEval-100 benchmark with 100 diverse scenes. As shown in~\cref{tab:sceneeval}, our method achieves strong improvements in relational consistency and physical plausibility, particularly in collision reduction and support relations. The runtime reported in~\cref{tab:sceneeval} excludes the rendering time. 

As shown in Tab. B, our method achieves improvements in relational consistency and physical plausibility, particularly in collision reduction and support relations, demonstrating good generalization beyond the original 33-scene benchmark. 

\begin{figure}[!tp]
\vspace{-2mm}
\centering
\includegraphics[width=0.9\linewidth]{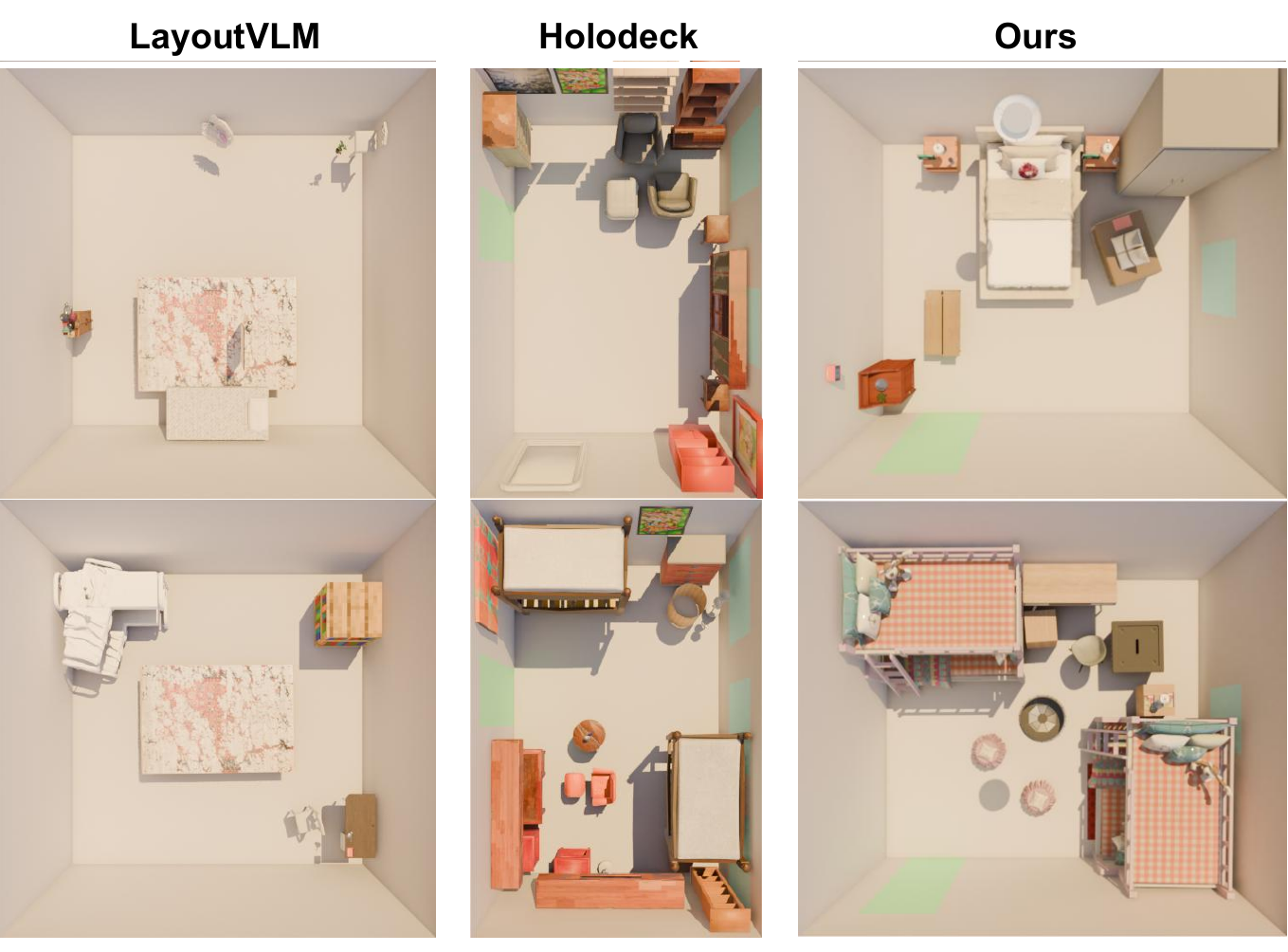}
\caption{\textbf{SceneEval-100 visualization.}}
\label{fig:dynamics}
\end{figure}
\section{VLM-Based Scene Layout Representation}
This section provides details about VLM-Based Scene Layout Representation, as referenced in Sec.~3 of the main paper.
\renewcommand{\thefigure}{\Alph{figure}}
\renewcommand{\thetable}{\Alph{table}}
\subsection{Hierarchical Placement}
Horizontal support planes are detected by clustering coplanar mesh regions, following the hierarchical support modeling principles of HSM\cite{Pun25}. Given a furniture mesh $\mathcal{M}$, we extract all horizontal planes using:
\begin{equation}
\mathcal{P}_{\text{cluster}}(\mathcal{M}),
\end{equation}

and denote the global floor plane as $H_0$. The final set of horizontal support planes is thus the union of the floor plane and all detected furniture planes:
\begin{equation}
\mathcal{H} = \{H_0\} \cup \mathcal{P}_{\text{cluster}}(\mathcal{M}).
\end{equation}
This hierarchical decomposition reduces computational complexity and ensures both semantic and physical plausibility in the generated layouts.
\begin{figure}[htp]
    \centering
    \includegraphics[width=0.8\textwidth]{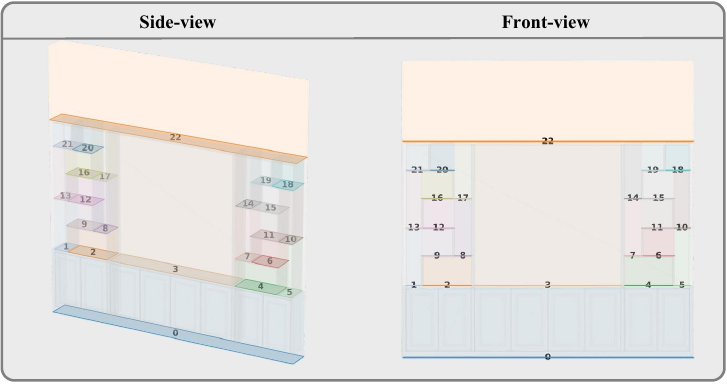}
    \vspace{-0.1cm}
    \caption{\textbf{Surface extraction results.} Once the support surfaces are detected, the scene’s side-view and front-view are produced and provided to the VLM as visual inputs for the following reasoning stage.}
    \label{fig:surface}
    \vspace{-0.4cm}
\end{figure}

\subsection{CLIP-based Asset Retrieval}

We employ a CLIP-based retrieval process to map these textual descriptions to actual 3D assets. Let the 3D asset database be:
\begin{equation}
\mathcal{A} = \{A_1, A_2, \dots, A_K\},
\end{equation}
and for each asset $a_j$, a corresponding feature embedding $f_{clip}^{I}(a_j)$ is computed using the CLIP image encoder, where the input image is the rendered front-view of $a_j$. For each textual query $s_i^m$, we compute the CLIP similarity:
\begin{equation}
C_{ij} = \langle f_{clip}^{T}(s_i^m), f_{clip}^{I}(a_j) \rangle,
\end{equation}
where $f_{clip}^{T}(\cdot)$ denotes the text encoder of CLIP. The top-\emph{k} candidate assets in terms of semantic relevance are selected first. For each candidate asset $a_j$, we further compute a bounding-box alignment error (MAE):
\begin{equation}
{E}_{ij} = \|b_i - b_j\|_1.
\end{equation}
This geometric term helps distinguish assets with similar semantics but different sizes or aspect ratios, which CLIP alone cannot differentiate.

We then combine the CLIP similarity $C_{ij}$ and alignment error $E_{ij}$ into a final matching score:
\begin{equation}
S_{ij} = w_1\, C_{ij} - w_2\, E_{ij},
\end{equation}
where $w_1$ and $w_2$ balance semantic similarity and geometric consistency. The asset with the highest $S_{ij}$, subject to size and category constraints, is selected as the matched instance $a_i$.

\subsection{Constraint Generation}
We define six types of constraints for floor-level objects and seven for furniture-level objects, categorized as shown in Tab.~\ref{tab:constraints}. For the specific formula, see~\ref{sec:semantic_supp}

\section{Constraints and Loss Functions}
\begin{table}[ht]
\vspace{-0.4cm}
\centering
\caption{Predefined Spatial Constraints for Floor-level and Furniture-level Assets}
\vspace{-0.2cm}
\begin{tabular}{p{2.4cm} p{3.7cm} p{4.6cm}}
\hline
\textbf{Asset Level} & \textbf{Constraint Category} & \textbf{Constraint Functions} \\
\hline

\multirow{6}{*}{Floor-level} 
& \multirow{2}{*}{Position based}
& $\text{distance}(P_i, P_j, d_{\min}, d_{\max})$ \\
&
& $\text{place\_align}(P_i, P_j, \text{dir})$ \\
\cline{2-3}

& \multirow{2}{*}{Orientation based}
& $\text{align\_with}(P_i, P_j)$ \\
&
& $\text{point\_towards}(P_i, P_j)$ \\
\cline{2-3}

& \multirow{2}{*}{Dual based}
& $\text{against\_wall}(P_i, wall_i)$ \\
&
& $\text{surround}(P_i, P_j, d)$ \\

\hline

\multirow{7}{*}{Furniture-level} 
& \multirow{3}{*}{Position based}
& $\text{center}(P_i, region_j)$ \\
&
& $\text{distance}(P_i, P_j, d_{\min}, d_{\max})$ \\
&
& $\text{place\_align}(P_i, P_j, \text{dir})$ \\
\cline{2-3}

& \multirow{2}{*}{Orientation based}
& $\text{align\_with}(P_i, P_j)$ \\
&
& $\text{point\_towards\_edge}(P_i, edge_i)$ \\
\cline{2-3}

& \multirow{2}{*}{Dual based}
& $\text{against\_edge}(P_i, edge_i)$ \\
&
& $\text{surround}(P_i, P_j, d)$ \\

\hline
\end{tabular}
\label{tab:constraints}
\vspace{-0.4cm}
\end{table}
This section summarizes the constraint formulations and loss functions used throughout Secs.~4.3 and~4.4.
These constraints encode geometric, semantic, and physical relations between scene assets, while the associated losses provide differentiable objectives that drive the optimization toward layout configurations consistent with these relations. Some formulations adopted from LayoutVLM\cite{Sun25}.

\subsection{Semantic Loss}
\label{sec:semantic_supp}
We provide the implementation details of the semantic loss functions used for layout refinement.
 Throughout this section, we use the following notation:

- $p_i=[x_i, y_i, z_i]\in\mathbb{R}^2$: the center position of asset $a_i$ on the ground floor;

- $v_i$: the 2D unit vector encoding the yaw of asset $a_i$, computed from its rotation angle $\theta_i$ around the $z$-axis as $v_i=(\cos\theta_i,\sin\theta_i)$.

- For walls $wall_i$, $p_i=[x_i, y_i, z_i]\in\mathbb{R}^2$ is the center bottom point of the wall and $v_i$ is its outward surface normal.

Each constraint is fully differentiable and designed to guide assets toward semantically plausible spatial relations.

\paragraph{Against-Wall.}
To place an asset $a_i$ flush against wall $wall_i$, the loss is decomposed into a distance term and an orientation term.

\noindent{(1) Distance terms.}
Let $g_i\in\mathbb{R}$ denote the half-thickness of asset $a_i$ measured along the wall normal direction. The distance penalty enforces the object–wall offset to match this thickness:
\begin{equation}
L_{\text{dist}}(a_i,wall_i)
=
\Bigl|\langle p_i - p_j,\; v_j\rangle - g_i\Bigr|.
\end{equation}

\noindent{(2) Orientation terms.}
Using the previously defined facing vector $v_i$, orientation consistency with the wall normal is enforced via:
\begin{equation}
L_{\text{ori}}(a_i,wall_i)
=
1 - v_i \cdot v_j.
\end{equation}

Combining the two components yields:
\begin{equation}
L_{\text{AgainstWall}}(a_i,wall_i)
=
L_{\text{dist}}(a_i,wall_i)
+
L_{\text{ori}}(a_i,wall_i).
\end{equation}

\paragraph{Near.}
To constrain the distance between two assets within $[d_{\min}, d_{\max}]$, we compute
the center distance or its wall-projected variant:
\begin{equation}
d_{ij} =
\begin{cases}
|\langle p_i - p_j,\; v_j\rangle|, & \text{if } a_j \text{ is a wall},\\[3pt]
\|p_i - p_j\|_2, & \text{otherwise}.
\end{cases}
\end{equation}
A bounded-distance loss is applied:
\begin{equation}
L_{\text{Near}}(a_i,a_j)
=
\operatorname{ReLU}(d_{\min}-d_{ij})
+
\operatorname{ReLU}(d_{ij}-d_{\max}),
\end{equation}
where the bounds are automatically expanded to account for asset bounding-box sizes.

\paragraph{Align-With.}
For rotation-consistency constraints, we follow a cosine-distance formulation.
Let $v_i=(\cos r_i,\sin r_i)$ and $v_j=(\cos (r_j+\phi), \sin (r_j+\phi))$.
The loss is:
\begin{equation}
L_{\text{Align}}(a_i,a_j,\phi)
=
1 - v_i \cdot v_j .
\end{equation}
This encourages the orientation of $a_i$ and $a_j$ to differ by exactly $\phi$.

\paragraph{Point-Towards.}
To orient object $a_i$ toward target $a_j$, we compare the object's facing direction $v_i$
with the desired direction $d_{ij}=p_j - p_i$:
\begin{equation}
L_{\text{PointTowards}}(a_i,a_j)
=
1 -
\frac{\langle d_{ij},\, v_i\rangle}{\|d_{ij}\|_2 + \varepsilon}.
\end{equation}

\paragraph{Surround.}
Given a center object $A_c$ and a set of surrounding objects $\{A_k\}_{k\in S}$, the loss consists of distance-related and orientation-related components.

\noindent{(1) Distance terms.}
We first encourage the surrounding objects to be distributed symmetrically around the center of the central object:
\begin{equation}
L_{\text{center}}
=
\left\|
\frac{1}{|S|}\sum_{k\in S} p_k - p_c
\right\|_2.
\end{equation}
This term measures the deviation of the centroid of the surrounding objects $S$ from the center position $p_c$ of the central object, promoting a balanced, roughly symmetric arrangement around it.
For each surrounding object, we further enforce appropriate proximity:
\begin{equation}
L_{\text{dist}}
=
\frac{1}{|S|}
\sum_{k\in S}
L_{\text{Near}}(A_k, A_c).
\end{equation}

\noindent{(2) Orientation term.}
 To make surrounding objects face the center, we apply:
\begin{equation}
L_{\text{ori}}
=
\frac{1}{|S|}
\sum_{k\in S}
L_{\text{PointTowards}}(A_k, A_c).
\end{equation}
Combining all components gives the full surround loss:
\begin{equation}
L_{\text{Surround}}
=
L_{\text{center}}
+
L_{\text{dist}}
+
L_{\text{ori}}.
\end{equation}
This decomposition separately captures radial distribution, proximity control, and center-facing orientation, yielding a balanced symmetric arrangement around $A_c$.

\paragraph{Place-Align.}
This constraint encourages asset $a_i$ to be placed in a semantic direction 
$d\in\{\text{up,down,left,right}\}$ relative to $a_j$.
we align the vector from $a_j$ to $a_i$ with the target direction $t_d$ and suppress perpendicular drift. Let
\begin{equation}
u_{ij}=\frac{p_i - p_j}{\|p_i - p_j\|_2},
\end{equation}
The target direction $t_d$ is a fixed world-axis unit vector determined by the semantic direction label $d$:
\begin{equation}
\small{
t_{\text{right}}=(1,0),\quad   
t_{\text{left}}=(-1,0),\quad
t_{\text{up}}=(0,1),\quad
t_{\text{down}}=(0,-1).}
\end{equation}
We additionally define $n_d$ as the axis perpendicular to $t_d$, used to suppress lateral drift.

The loss combines directional alignment and perpendicular deviation:
The loss is
\begin{equation}
L_{\text{PlaceAlign}}(a_i,a_j,d)=\lambda_1\bigl(1 - u_{ij}\cdot t_d\bigr)+\lambda_2\,\bigl|\langle p_i - p_j,\; n_d\rangle\bigr|.
\end{equation}

The following constraints are newly introduced for furniture-level assets

\paragraph{Center.}

To pull an asset $a_i$ toward the geometric center of a region $\mathcal{R}$, let
\begin{equation}
p_c = \left(\frac{x_{\min}+x_{\max}}{2},\;\frac{y_{\min}+y_{\max}}{2}\right)
\end{equation}
denote the region center.
We minimize the L2 distance:
\begin{equation}
L_{\text{Center}}(a_i,\mathcal{R})
=
\|\, p_i - p_c \,\|_2.
\end{equation}

\paragraph{Against-Edge.}

For an object $a_i$ placed on a supporting furniture surface and a boundary edge $e$, the loss is decomposed into a distance term and an orientation term.

\noindent{(1) Distance term.}
Let $q_e$ be the projection point on edge $e$, and $n_e$ its outward normal.
We only measure how far the object sits relative to the edge:
\begin{equation}
L_{\text{dist}}(a_i,e)
=
\bigl|\langle p_i - q_e,\; n_e\rangle - g_i \bigr|.
\end{equation}
Here, the boundary edge $e$ refers to an edge of the supporting furniture’s top surface.

\noindent{(2) Orientation term.}
Let $t_e$ be the desired facing direction associated with edge $e$.
We penalize angular disagreement via cosine:
\begin{equation}
L_{\text{ori}}(a_i,e)= 1 - v_i\cdot t_e.
\end{equation}

Combining all components gives the full loss:
\begin{equation}
L_{\text{AgainstEdge}}(a_i,e)
=
L_{\text{dist}}(a_i,e)
+
L_{\text{ori}}(a_i,e).
\end{equation}

\paragraph{Point-Towards-Edge.}

To orient object $a_i$ toward a region edge $e$, let $q_e$ denote the closest point on that edge.
 The desired pointing direction is:
\begin{equation}
d_{ie} = q_e - p_i.
\end{equation}
The angle loss is computed as:
\begin{equation}
L_{\text{PointTowardsEdge}}(a_i,e)
=
1 - 
\frac{\langle d_{ie},\, v_i\rangle}
{\|d_{ie}\|_2 + \varepsilon}.
\end{equation}
This encourages the object to face toward the target edge.

\subsection{Physical Loss}

We employ two differentiable physical validity terms:
(i) a collision-avoidance loss based on oriented bounding boxes (OBBs), and
(ii) an in-bound constraint that ensures all assets remain within the room boundary.

\noindent\textbf{Collision Loss.}
We employ a differentiable 2D oriented bounding box (OBB) collision loss based on the Separating Axis Theorem (SAT)\cite{Boyd04}. For any two OBBs $A$ and $B$, we construct four candidate separating axes,
\begin{equation}
\mathcal{U}=\{u_1^A, u_2^A, u_1^B, u_2^B\},
\end{equation}
corresponding to the edge normals of both boxes. Each OBB is represented by its four corner points $\{p_k\}_{k=1}^{4}$, which are projected onto each axis $u_i$ to form the scalar interval
\begin{equation}
[n_i, m_i] = 
\left[
\min_k \langle p_k, u_i \rangle,
\max_k \langle p_k, u_i \rangle
\right].
\end{equation}
For axis $u_i$, the interval overlap is defined as a ReLU:
\begin{equation}
o_i = \mathrm{ReLU}\!\left(
\min(m_i^A,m_i^B) - \max(n_i^A,n_i^B)
\right),
\end{equation}
where $o_i>0$ indicates overlap on that axis. To obtain a stable, rotation-sensitive collision loss, we aggregate the overlaps using a harmonic mean:
\begin{equation}
L_{\mathrm{col}}(A,B) 
= 
\frac{4}
{\sum_{i=1}^{4} \frac{1}{o_i + \varepsilon}} ,
\end{equation}
where $\varepsilon$ is a small constant. 
This formulation offers three advantages:
(i) strong gradients whenever any axis becomes separating, which encourages non-intersection;
(ii) smooth behavior under both translation and rotation; and
(iii) improved numerical stability compared with directly using $\min(o_i)$.
The total physical loss is computed by summing $L_{\mathrm{OBB}}$ over all object pairs.

\begin{figure}[t]
\centering
\includegraphics[width=0.95\linewidth]{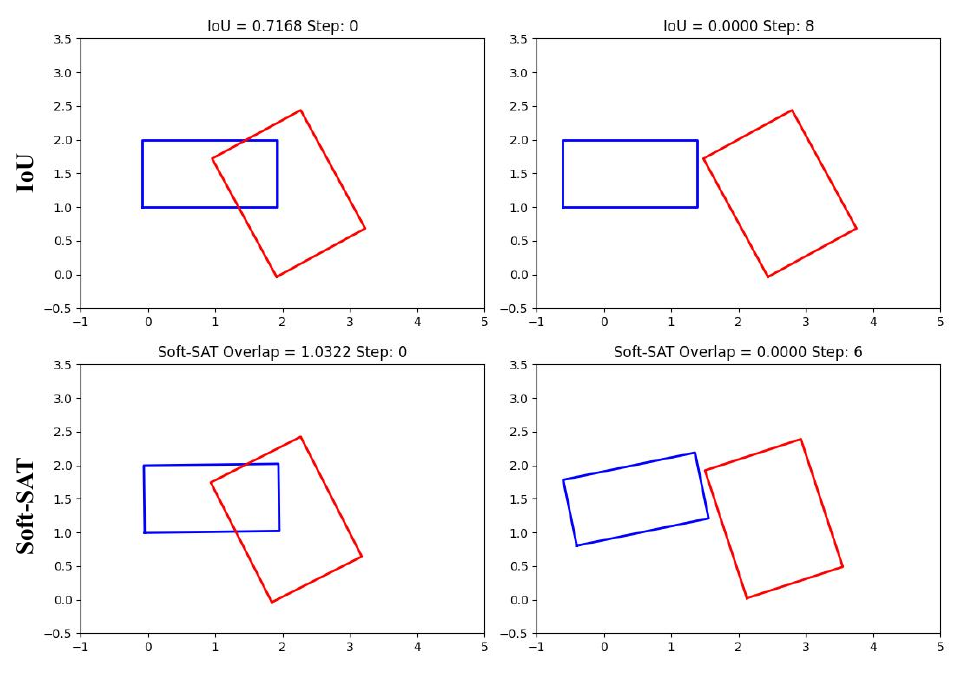}
\caption{\textbf{Collision loss.} Comparison of optimization trajectories using IoU loss (top) and our Soft-SAT loss (bottom).}
\vspace{-0.55cm}
\label{fig:loss}
\end{figure}

As shown in \cref{fig:loss}, starting from the same initial overlapping configuration, we performed 10 gradient descent iterations with a learning rate of 0.1. The Intersection-over-Union (IoU) loss typically achieves only translation-based separation. In contrast, Soft-SAT maintains differentiability near contact boundaries, continuously providing meaningful gradients. This enables faster convergence and supports rotation-driven decoupling.

\noindent\textbf{In-Bound Loss.}
To enforce spatial validity, each asset is required to remain inside the room's boundary polygon~$B$.
For each corner point $p_k$ of an OBB, let $d(p_k)$ denote its signed distance to the boundary,
where positive values indicate points inside the boundary and negative values indicate points outside.
We penalize any boundary violation using
\begin{equation}
L_{\mathrm{in}}(A)
=
\frac{1}{4}
\sum_{k=1}^{4}
\mathrm{ReLU}\!\bigl(-d(p_k)\bigr).
\end{equation}
The total physical loss combines collisions and boundary validity:
\begin{equation}
L_{\mathrm{physics}}
=
\sum_{A<B} L_{\mathrm{col}}(A,B)
+
\sum_{A} L_{\mathrm{in}}(A).
\end{equation}

\section{Global Semantic Verification}

\subsection{Subgraph Segmentation and Central Node Selection.}
We first partition $G_0$ into multiple connected components
\begin{equation}
\mathcal{G} = \{ G^1, G^2, \dots, G^L \},
\end{equation}
where each subgraph $G^l = (V^l, E^l)$ corresponds to a functional area or a localized object group.
Within each subgraph \(G^l\), we compute the weighted centrality of each node \(v_i^l \in V^l\) as:
\begin{equation}
C_{v_i^l} = c_l(v_i^l) \cdot |V^l| = 
\frac{\deg(v_i^l) \cdot |V^l|}{|V^l| - 1},
\end{equation}
where \(\deg(v_i^l)\) denotes the in-degree of node \(v_i^l\), 
\(c_l(v_i^l)\) represents the normalized centrality of node \(v_i^l\) within subgraph \(G^l\), 
and \(|V^l|\) is the number of nodes contained in the connected component \(G^l\). To prevent the relative importance of nodes in larger groups from being diminished, the term is scaled by the group size \(|V^l|\).  

We designate the node with the highest \(C_{v_i^l}\) as the center node, 
indicating the object with the greatest influence on the local layout. 
When multiple nodes satisfy the condition \(C_{v_i^l} > \tau\), that is,
\begin{equation}
|\{ v_i^l \in V^l \mid C_{v_i^l} > \tau \}| > 1,
\end{equation}
the subgraph \(G^l\) is further split into smaller subgraphs:
\begin{equation}
G^l \rightarrow \{ G^{l_1}, G^{l_2}, \dots, G^{l_m} \},
\end{equation}
ensuring that each resulting subgraph contains exactly one dominant center object, which serves as the key anchor for subsequent conflict detection and group-wise mutation during evolutionary optimization.

This section expands the consistency validation step in Sec.~3.1.2 by detailing how each conflict term
$\Phi=\{\Phi_{\text{cycle}},\, \Phi_{\text{dist}},\, \Phi_{\text{wall}},\, \Phi_{\text{sem}},\, \Phi_{\text{comp}},\, \Phi_{\text{oa}}\}$
is evaluated on the constraint graph.
Each $\Phi$ corresponds to a distinct family of violations derived from the graph structure, 
wall and metric relations, or semantic rules attached to asset pairs.

\noindent\textbf{(1) Directional Cycle Conflict—$\Phi_\text{cycle}$}

A directional cycle conflict occurs if there exists a closed loop among alignment-type edges. Formally,
\begin{equation}
\exists~\text{cycle}~\mathcal{P} = (v_1,v_2,\dots,v_n,v_1),
\end{equation}
where all edges in the cycle are of type place align.

\noindent\textbf{(2) Distance Inconsistency—$\Phi_\text{dist}$}

A distance inconsistency arises when the actual path length between two nodes
deviates from the specified target distance. Formally,
\begin{equation}
|\text{path\_len}(v_i,v_j) - d^*_{ij}| > \epsilon_d.
\end{equation}
This condition applies to edges of type Near.

\noindent\textbf{(3) Mutual-exclusive Wall Conflicts—$\Phi_\text{wall}$}

\noindent\textbf{Multiple Against-Wall.} A node must not be constrained against more than one wall:
\begin{equation}
|\{\, w : (v_i,w) \in \mathcal{A} \,\}| > 1
\end{equation}

\noindent\textbf{Mixed Align/Against.} A node cannot simultaneously have an align-wall and an against-wall constraint with different wall:
\begin{equation}
\exists\, (v_i,w_1)\in\mathcal{L},\;
      (v_i,w_2)\in\mathcal{A}
\end{equation}

\noindent\textbf{Wall Over-Occupancy.}
The total projected length of all objects attached to a wall must not exceed its capacity:
\begin{equation}
\sum_{(v_i,w)\in \mathcal{A}} L(v_i)
\;>\;
L_{\text{bound}}(w)
\end{equation}

\noindent\textbf{(4) Out-of-Area Violation—$\Phi_\text{oa}$}
 
Each asset must lie entirely inside the valid layout region.  
Let $p_k^i$ denote the four OBB corner points of node $v_i$, and let $d(p_k^i)$ be the signed distance to the room boundary or the corresponding support region boundary (positive inside).  
For small-object layouts on an open support region $r$, we also consider the region invalid if its against-edge constraints leave insufficient interior feasible space. 
An out-of-area conflict occurs if
\begin{equation}
\exists\, k:\; d(p_k^i) < 0
\quad \lor \quad
\Omega_r(\mathcal{E}_r) < \epsilon,
\end{equation}
where $\mathcal{E}_r=\{(v_i,e)\in\mathcal{E}: e \subset \partial r\}$ denotes the set of \texttt{against\_edge} constraints associated with the boundary of region $r$, and $\Omega_r(\mathcal{E}_r)$ measures the remaining feasible interior space after applying these edge constraints. 
The first condition indicates that an asset leaks outside the valid region, while the second captures over-constrained open support regions with insufficient placement space.

All detected violations are stored in the unified conflict log $L_{\text{log}}$ together with their corresponding edges and constraint types.

\noindent\textbf{(5) Logic Violation—$\Phi_\text{logic}$}
 
\noindent\textbf{Center–Edge Conflict.} An asset centered on a region must not simultaneously be constrained against one of its edges.
Formally,
\begin{equation}
\exists\,(v_i,r)\in\mathcal{C},\;
      (v_i,e)\in\mathcal{E},
\end{equation}
where $\mathcal{C}$ denotes center constraints and $\mathcal{E}$ denotes against-edge constraints.

\noindent\textbf{Region Over-Centering.} Each region can host at most one centered asset.  
A violation occurs if multiple nodes are assigned to the same region:
\begin{equation}
|\{\, v_i : (v_i,r)\in\mathcal{C} \,\}| > 1 .
\end{equation}

\noindent\textbf{(6) Outgoing Edge Completeness—$\Phi_\text{comp}$}

Each node $v_i$ must have both position- and orientation-type outgoing edges.
Let

- $\mathcal{C}_{\text{pos}}$: set of position-type constraints

- $\mathcal{C}_{\text{ori}}$: set of orientation-type constraints

A node is incomplete if
\begin{equation}
\neg~\exists\, e_{ij}\in\mathcal{C}_{\text{pos}}
\quad \lor \quad
\neg~\exists\, e_{ij}\in\mathcal{C}_{\text{ori}},
\end{equation}
indicating missing spatial constraints that will be corrected in the next refinement stage.

\noindent\textbf{(7) Semantic Consistency Violation—$\Phi_\text{sem}$}

Each semantic group $g$ specifies
(i) a set of permitted relation types $R_g$, and 
(ii) admissible attribute ranges $\Omega_g$ for those relations.
A semantic conflict is triggered if either
\begin{equation}
(v_a, v_b) \notin R_g
\quad\text{or}\quad
\mathrm{arg}(v_a, v_b) \notin \Omega_g .
\end{equation}

\section{Additional ablation results}
\begin{figure}[htp]
    \centering
    \includegraphics[width=\textwidth]{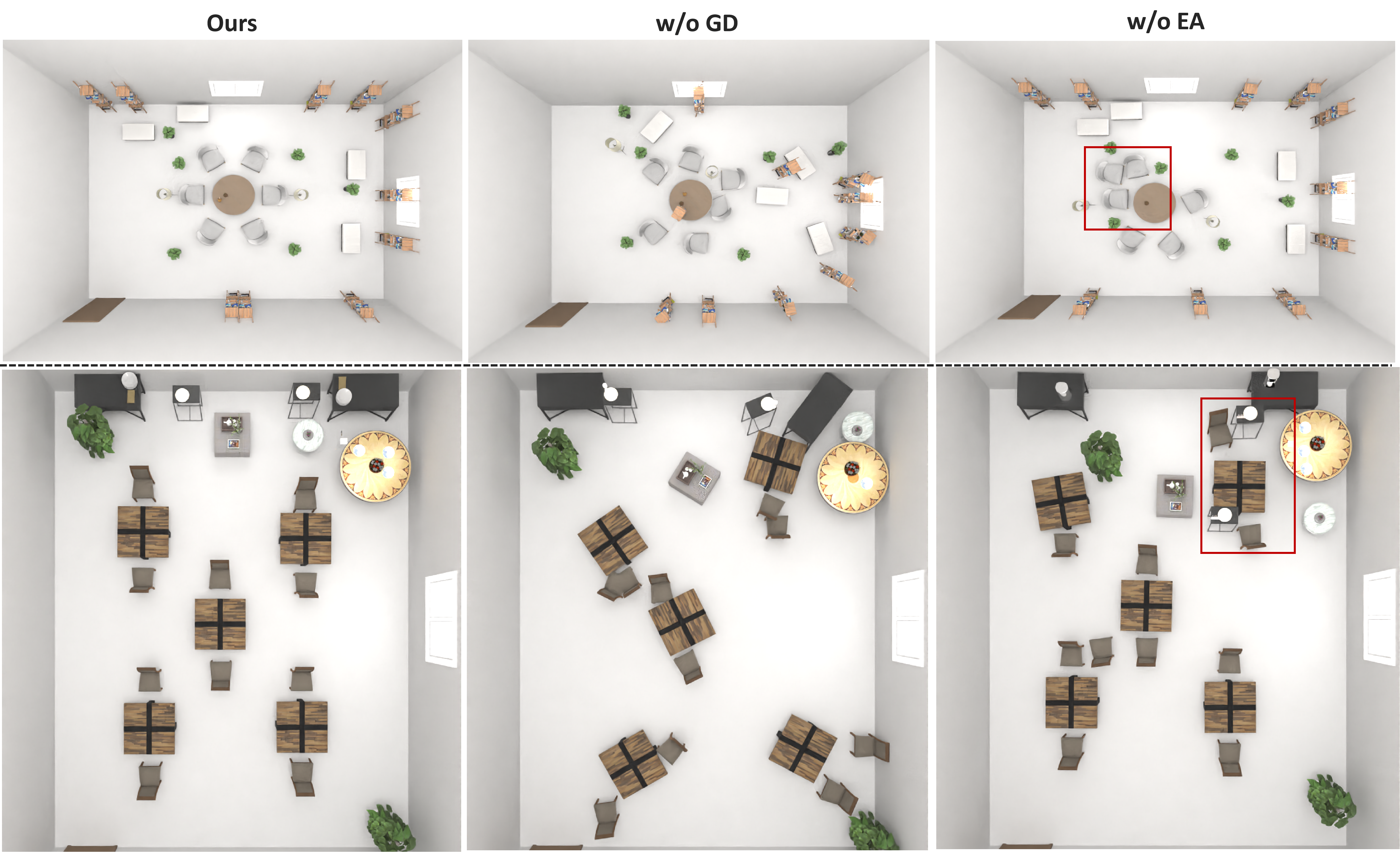}
    \caption{\textbf{Additional ablation results at floor level.} Comparison of layout results with or without EA and Assets overlap at the centerGD module.}
    \label{fig:ablation}
\end{figure}
\begin{figure}[htp]
    \centering
    \includegraphics[width=0.7\textwidth]{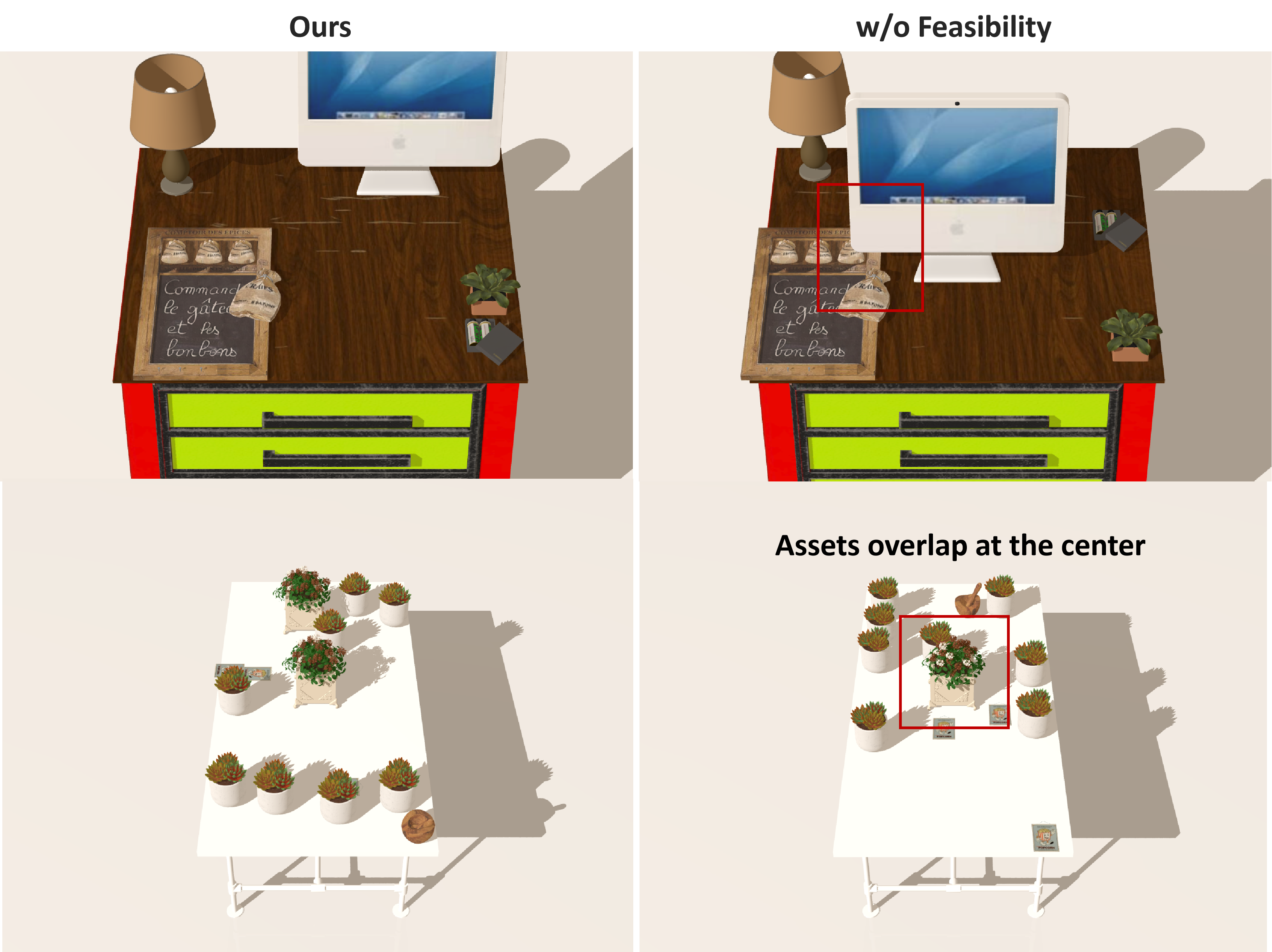}
    \caption{\textbf{Additional ablation results at furiture level.} Comparison of layout results with or without Feasibility and Completeness at furiture level placement.}
    \label{fig:ablation}
\end{figure}
We provide additional ablation results to complement the analysis in the main paper. Not all GSV submodules are applied in every scene; they are activated only when the corresponding constraint conflicts arise. As shown in Fig. 2, removing EA leads to less organized layouts and increased object overlap, while disabling GD results in local misplacements. At the furniture level (Fig. 3), removing the feasibility constraint causes overlapping or physically implausible placements. Together with the results presented in the main paper, these observations further demonstrate the complementary roles of these components in producing coherent and physically valid layouts.

\section{Details of Prompt}
In this section, we provide all prompts used throughout our scene generation pipeline. The pipeline consists of seven main stages, ranging from initial floor plan generation to the placement of small assets on large furniture.

\subsection{Preliminary Floor Plan Generation}
The first stage generates an initial floor plan from textual instructions, including the positions of doors and windows, adapted from the Holodeck\cite{Yang24} framework. The output serves as the foundational geometric layout for subsequent steps. 
\UseRawInputEncoding{
\styledfileinput[python]{program/floor-plan.txt}{}
}

\subsection{Functional Area Segmentation}
Once a preliminary floor plan is available, we segment the space into functional areas. This step takes as input: (i) the floor plan from Step~1, (ii) textual layout instructions, and (iii) optional asset categories. If asset categories are not provided, the VLM predicts suitable objects for each area. Otherwise, the layout generation adheres to the specified categories. 
\UseRawInputEncoding{
\styledfileinput[python]{program/get-areas.txt}{}
}

\subsection{Asset Placement per Functional Area}
With functional areas defined, the VLM populates each area with objects, providing both semantic descriptions and approximate sizes for each asset. This step bridges the abstract functional division with concrete object-level placement.
\styledfileinput[python]{program/get-objects.txt}{}

\subsection{Constraint Generation}
After acquiring all assets, we input the following to the VLM: (i) the layout instruction $\ell_{\text{layout}}$, which may range from brief prompts to detailed directives; (ii) the current 3D scene state $S_t$, represented as a 2D floor plan with placed objects, doors, and windows; and (iii) a rendered single-object view of each asset $a_i$. The VLM then generates per-object constraints, which are output as executable Python code. These constraints serve as differentiable objectives in subsequent optimization.
\UseRawInputEncoding{
\styledfileinput[python]{program/constraints.txt}{}
}

\subsection{Large Furniture Identification}
Upon completing floor-level asset placement, the rendered scene and textual prompts are fed to the VLM to identify large furniture items that can be added to the layout. This step ensures that the overall arrangement respects the spatial requirements of major assets.
\UseRawInputEncoding{
\styledfileinput[python]{program/find-big-objects.txt}{}
}

\subsection{Small Asset Placement}
We first cluster mesh triangles to extract all potential horizontal support surfaces. The resulting side-view and main-view renderings of these surfaces (shown in Fig.~\ref{fig:surface}), together with the large-furniture descriptions, reference images, and the overall rendered layout, are then provided to the VLM. Using these multimodal inputs, the VLM plans the placement of small assets and determines which objects should be supported by each piece of furniture, thereby populating the scene with detailed items while respecting physical support constraints.
\UseRawInputEncoding{
\styledfileinput[python]{program/small-assets.txt}{}
}

\subsection{Constraint Assignment for Small Assets}
Finally, the small asset information is input to the VLM, which generates per-object constraints for each small asset. If no constraints are necessary for a given object, an empty code block is returned. This ensures that even minor objects are appropriately constrained within the scene for semantic plausibility and physical consistency.
\UseRawInputEncoding{
\styledfileinput[python]{program/small-constraints.txt}{}
}

\subsection{Conflict Feedback Prompts}
Building on the conflicts identified in Sec~3.1.3, all detected conflicts ($\Gamma(e_{ij})=1$) are recorded in a unified conflict log $L_\text{log}$, along with their associated edges and constraint types. This collected conflict log is then fed to the VLM, which interprets the logged conflicts and proposes adjustments to resolve structural or semantic inconsistencies. By iteratively providing such conflict logs as prompts, the pipeline refines the layout to ensure semantic coherence and physical plausibility.
\UseRawInputEncoding{
\styledfileinput[python]{program/conflict-reflection.txt}{}
}

\end{document}